\pdfoutput=1

\documentclass{article}
\usepackage{iclr2027_conference}
\usepackage[T1]{fontenc}
\renewcommand{\sfdefault}{phv}

\usepackage{amsmath,amsfonts,bm}

\def\eqref#1{equation~\ref{#1}}

\def\1{\bm{1}}

\DeclareMathAlphabet{\mathsfit}{\encodingdefault}{\sfdefault}{m}{sl}
\SetMathAlphabet{\mathsfit}{bold}{\encodingdefault}{\sfdefault}{bx}{n}

\usepackage{hyperref}
\hypersetup{hidelinks,pdfauthor={Miaobo Hu, Shuhao Hu, Xiaobo Guo, Xin Wang, Bokun Wang, Tianshu Fu, Daren Zha},pdftitle={RouteAudit: Interaction-Aware Identification for Budgeted Multi-Verifier Routing}}
\usepackage{url}
\usepackage{graphicx}
\usepackage{booktabs}
\usepackage{tabularx}
\usepackage{array}
\usepackage{needspace}
\newcolumntype{L}{>{\raggedright\arraybackslash}X}
\usepackage{microtype}
\usepackage{xcolor}
\definecolor{raBlue}{HTML}{0072B2}
\definecolor{raOrange}{HTML}{D55E00}
\definecolor{raGreen}{HTML}{009E73}
\definecolor{raGray}{HTML}{6B7280}
\definecolor{raPurple}{HTML}{CC79A7}

\title{ROUTEAUDIT: Interaction-Aware\\Identification for Budgeted\\Multi-Verifier Routing}
\author{Miaobo Hu$^{1,2,*}$, Shuhao Hu$^{2}$, Xiaobo Guo$^{2}$, Xin Wang$^{2}$,\\
Bokun Wang$^{2}$, Tianshu Fu$^{2}$, Daren Zha$^{2}$\\[4pt]
{\normalfont $^{1}$School of Artificial Intelligence, University of Chinese Academy of Sciences, Beijing, China}\\
{\normalfont $^{2}$Institute of Information Engineering, Chinese Academy of Sciences, Beijing, China}\\
{\normalfont $^{*}$Corresponding author: \texttt{humiaobo@iie.ac.cn}}}

\iclrfinalcopy
\begin{document}
\maketitle
\lhead{Preprint}

\begin{abstract}
Adaptive multi-verifier systems are commonly compared through endpoint quality--cost gaps, even
when the verifier catalog, availability, accounting, information filtration, or scorer changes
with the policy. We formulate verifier routing as a contract-conditioned identification problem.
The contract records request support, verifier catalog, realized availability, resource
accounting, online filtration, and post-trace scoring; a matched route contrast changes only the
policy coordinate.

ROUTEAUDIT adds three measurable objects to this contract. A contract lattice averages coordinate
increments over every admissible bridge order and reports the resulting attribution together with
its path sensitivity. A policy-independent response tape identifies paired sequential contrasts
when adaptive policies reveal different observations. For incomplete matching, request-level
bounds use whichever potential outcome remains observed and give a sharp finite-population
interval.

The protocol commits paid observations and ledger events before the oracle join and returns an
attribution certificate for each comparison. On two held-out raw-tail caches, matched static
SF+SA equals the cascade, assigning the apparent gains of 0.1797 and 0.1250 over full static to
the verifier-set edge. On 1,319 held-out task requests, the learned and RLVR studies
report quality 0.9522 and 0.9553 versus 0.9484 for matched static; the RLVR--static paired
difference is $+0.0068$ with a request-paired interval $[0.0015,0.0122]$ and a
training-seed-by-request hierarchical interval $[0.0006,0.0131]$. Controlled attribution
recovery yields route mean absolute error 0.0011 and endpoint reconstruction error 0.0004.
Factorial, bridge-order, and stochastic-provider studies evaluate the certificate interface;
RLVR supplies a learned-policy stress test under the same identification contract.
\end{abstract}

\section{Introduction}
A cheap verifier followed by a conditional expensive verifier is an attractive way to spend a finite inference budget. A rigorous comparison keeps the verifier set, cache, budget, and information boundary visible because a static policy and a cascade can change several of these objects at once. RouteAudit turns that requirement into three estimands and a trace contract that can be checked before the final oracle join.

We make attribution, rather than routing accuracy alone, the main object of study. Let
$\mathcal{K}=(\mathcal{I},\mathcal{V},\mathcal{A},\mathcal{P},\mathcal{F},\mathcal{S})$ denote the
ordered request support, verifier catalog, realized availability schedule, prices and resource
accounting, online information filtration, and post-trace scorer. For policy $\pi$, write
$Y(\pi;\mathcal{K})$ for the reported outcome. The naive contrast
\[
\Delta_{\mathrm{naive}} =
Y(\pi_{\mathrm{cascade}};\mathcal{K}_{\mathrm{cas}})
-
Y(\pi_{\mathrm{static}};\mathcal{K}_{\mathrm{sta}})
\]
mixes a route change with any changed contract coordinate. A route claim requires the matched
contrast
\[
\Delta_{\mathrm{route}} =
Y(\pi_{\mathrm{cascade}};\mathcal{K})
-
Y(\pi_{\mathrm{static}};\mathcal{K}),
\]
which changes only the policy coordinate. The complementary verifier-set contrast is
\[
\Delta_{\mathrm{set}} =
Y(\pi_{\mathrm{static}};\mathcal{K}_{\mathrm{cas}})
-
Y(\pi_{\mathrm{static}};\mathcal{K}_{\mathrm{sta}}).
\]
The route effect can itself depend on the contract. If two contracts differ only in coordinate
$j$, define
\[
\Gamma_j = \Delta_{\mathrm{route}}(\mathcal{K}^{j=1})
           -\Delta_{\mathrm{route}}(\mathcal{K}^{j=0}).
\]
A nonzero $\Gamma_j$ means that the measured policy advantage changes with verifier access,
availability, accounting, filtration, or scoring. The two-verifier decomposition is therefore a
special case of an interaction-aware contract graph. When intermediate bridge nodes are measured,
an endpoint difference telescopes along each predeclared path; the coordinate attribution may vary
with bridge order and is summarized by path sensitivity. When only a known fraction of requests is
unmatched, bounded scores give a partial-identification interval instead of forcing a binary
identified/non-identified label.

The same contract applies to learned routing. A policy may use task outcomes from completed
training episodes to update its parameters, while the outcome for the current request remains
outside the online state. This separates the measurement question from the optimization
question: RouteAudit identifies which contrast is being measured, and RLVR supplies a delayed
reward policy evaluated under that contrast.

RouteAudit instantiates this audit on a budgeted multi-verifier case study. Historical cache studies
provide matched controls through separate small-format (SF), small--large agreement (SA), and
large-format (LF) checks. The task and serving studies apply the same information boundary to
fixed, calibrated, learned, and RLVR policies. Together, these settings expose when an apparent
gain comes from verifier choice and when a route achieves a better measured quality--cost trade-off.

Our contributions are:

\begin{enumerate}
\item \textbf{Contract-lattice identification and order-robust attribution.} We make request
support, verifier access, availability, accounting, filtration, and scoring explicit coordinates.
Permutation-averaged bridge increments decompose an endpoint change, expose pairwise interactions,
and retain $\Omega_j$ when an explanation depends on intervention order.
\item \textbf{Sequential identification under policy-dependent observations.} A response tape
indexed by request, verifier, and call index couples adaptive policies to the same provider
realization, separating a conditional paired estimand from independently resampled provider draws.
\item \textbf{Sharp partial identification with incomplete matching.} Request-level bounds use the
observed arm and the declared score range, strictly refining complete-case analysis when one arm
remains available.
\item \textbf{Machine-verifiable identification certificates.} Each comparison binds its contract,
matched support, closed traces, scorer, coupling regime, bridge set, and point estimate or interval;
the verifier returns \textsc{identified}, \textsc{partially-identified}, or \textsc{not-identified}.
\end{enumerate}

\section{Related Work}
Selective prediction frames confidence-based abstention as a coverage--risk trade-off \citep{geifman2019selectivenet}; FrugalGPT and RouteLLM make model-call quality--cost routing explicit \citep{chen2023frugalgpt,ong2024routellm}. Process verification distinguishes step-level evidence from a final answer label \citep{lightman2023verify,wang2023mathshepherd,processbench2024,cobbe2021gsm8k,qwenmath2024,shao2024deepseekmath}, while reward-model audits and reward-hacking analyses motivate keeping proxy scores separate from task outcomes \citep{lambert2024rewardbench,skalse2022rewardhacking}. Tool-use benchmarks show that a route must be evaluated against executable state and downstream tests rather than a local score \citep{yao2023react,schick2023toolformer,zhou2024webarena,jimenez2024swebench,liu2024agentbench,lu2024toolsandbox,yao2024taubench,debenedetti2024agentdojo}. Offline RL work motivates support-aware evaluation under distribution shift \citep{kumar2020cql,kostrikov2022iql,fu2021d4rl}, and self-feedback methods illustrate why feedback visibility must be recorded at the interface \citep{madaan2023selfrefine,shinn2023reflexion}.

\paragraph{Current routing frontier.}
Recent work makes the comparison boundary sharper. Adaptive Generate--Rank--Verify (ADAP) studies active generation and costly verification with explicit monotonicity assumptions and lower bounds \citep{adap2026}; CAPS allocates verifier compute along evidence and distribution axes \citep{caps2026}; SLARouter learns cost-optimal routing from sparse feedback with service-level guarantees \citep{slarouter2026}. Reward Model Routing uses offline reward-model strengths with online Bayesian selection \citep{rewardrouting2026}, and recent panel and test-time-compute work makes stopping, specialist selection, and introspection explicit \citep{panels2026,ziprc2026}. Process-Verified RL via Lean and AgentV-RL connect verifier feedback to policy optimization \citep{leanrl2026,agentv2026}, while Resample or Reroute studies recoverable stopping debt and the identification of subsequent actions \citep{resamplereroute2026}. RouteAudit complements these optimization objectives by testing whether an apparent cascade gain survives a matched verifier-set control, and by retaining that control in learned-policy and serving comparisons.

\paragraph{Positioning.}
The closest baselines share the accept-or-escalate shape, while RouteAudit makes the comparison
contract itself the object of measurement. Its technical object is an identification framework
for routing under changing information sets: paid-read traces, budget conservation, matched
verifier-set controls, route-diversity accounting, and a post-trace oracle join are recorded as
conditions on an estimand. Appendix~\ref{app:evaluation-protocols} instantiates the same contract
for learned routing, task scoring, and RLVR.

\begin{table}[t]
\caption{Mechanism-level comparison. ``Trace'' means that per-call observations and budget events are retained; ``matched'' means that a static policy uses the same verifier set and split as the adaptive policy. The final row is the RouteAudit contract used in this paper.}
\label{tab:mechanism-comparison}
\centering
\scriptsize
\setlength{\tabcolsep}{2.5pt}
\begin{tabularx}{\linewidth}{lXXXXXX}
\toprule
\textbf{family} & \textbf{outcomes} & \textbf{gate} & \textbf{budget trace} & \textbf{matched set} & \textbf{post-trace audit} & \textbf{task/live path}\\
\midrule
Selective prediction & task outputs & learned & aggregate & optional & risk--coverage & task\\
LLM routing & frozen or live outputs & learned & cost summary & benchmark dependent & aggregate & task\\
Process verification & step labels & verifier policy & verifier calls & verifier dependent & oracle join & task\\
Offline RL evaluation & trajectories & policy & support/cost & dataset dependent & replay & held-out task\\
RouteAudit & frozen traces & fixed or learned & per-call ledger & required & replay + mutation & task + serving\\
\bottomrule
\end{tabularx}
\end{table}

Table~\ref{tab:mechanism-comparison} positions the comparison contract alongside established routing and verification objectives. The experiments report the route and verifier-set contrasts, then apply the same matched comparison to task quality, serving latency, and learned policies.

\section{Method}
\paragraph{Task and output.}
A request comprises a prompt, candidate handles, verifier access, and a resource budget.
The route returns an answer, an accept/abstain decision, and an ordered paid-observation trace.
The normalized task scorer compares the terminal answer with the canonical target; the cache
proxy scores a verifier decision. The two targets retain separate denominators and names.
Candidate construction and system identities are recorded in Table~\ref{tab:experimental-system}.
\subsection{Observed-only protocol}
The online policy receives only the fields listed in the public audit contract: verifier ID, label, confidence, remaining budget, reason, observation availability, and measured local elapsed time. Oracle labels, target answers, hidden cards, raw generations, and future verifier outputs are stored separately. The trace is frozen before the oracle join. A run manifest records source-relative paths, configuration, model/cache hashes, source revision and dirty flag, and a deterministic replay digest. The public bundle removes absolute paths, host names, user names, checkpoints, and weights.

\begin{figure}[!t]
\centering
\includegraphics[width=\columnwidth]{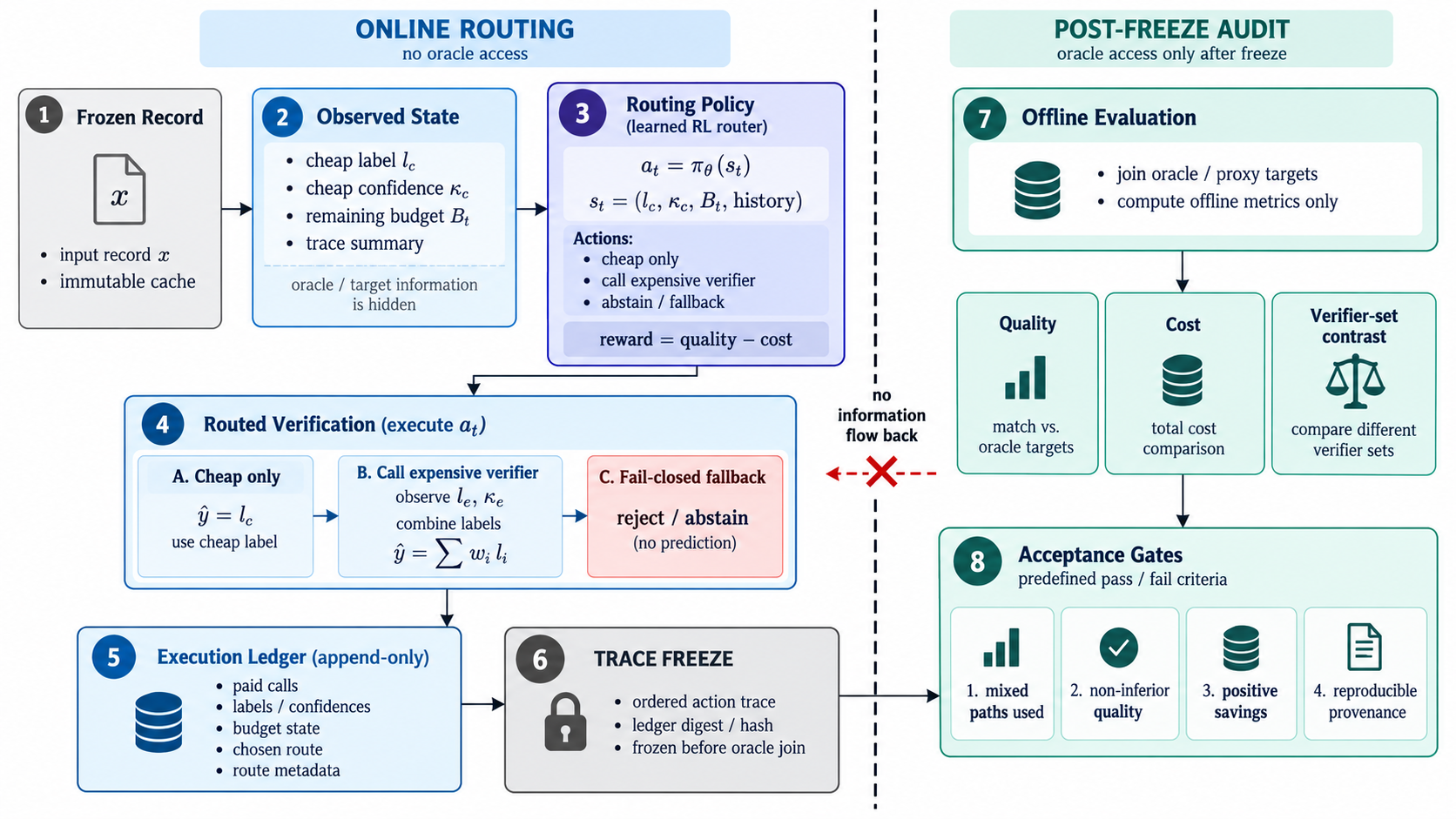}
\caption{RouteAudit's identification workflow. The online policy observes only the frozen record,
paid verifier outputs, and the remaining ledger; the trace is frozen before the post-freeze oracle
join. The audit then reports quality, cost, verifier-set contrasts, and acceptance gates from the
same committed trace.}
\label{fig:ra-protocol}
\end{figure}

\subsection{Estimands and matched controls}
For each example $i$, $C$ fixes the sample split, cache identity, availability policy, and ledger prices. The policy $\pi$ maps the allowed online trace to a sequence of verifier calls and a final decision. We report $Y$ as proxy accuracy when the source is a cache diagnostic, and separately report calls and declared units. The three contrasts are identifiable only when the same $C$ and the same verifier observations are available to both policies:
\begin{align}
\Delta_{\mathrm{naive}} &= Y(\pi_{\mathrm{cas}},V_{\mathrm{cas}},C)-Y(\pi_{\mathrm{sta}},V_{\mathrm{sta}},C),\\
\Delta_{\mathrm{route}} &= Y(\pi_{\mathrm{cas}},V,C)-Y(\pi_{\mathrm{sta}},V,C),\\
\Delta_{\mathrm{set}} &= Y(\pi_{\mathrm{sta}},V_{\mathrm{cas}},C)-Y(\pi_{\mathrm{sta}},V_{\mathrm{sta}},C).
\end{align}
The raw-tail track supplies the needed control: full static SF+SA+LF versus cascade is the naive comparison, while static SF+SA is the matched route control. If the cascade and static SF+SA have the same output on every record, then the observed full-static gap is attributable to the extra LF verifier under this cache and ledger. Algebraically, $\Delta_{\mathrm{naive}}=\Delta_{\mathrm{route}}+\Delta_{\mathrm{set}}$ when $V=V_{\mathrm{cas}}$.

\subsection{Identification and interaction contract}
\label{sec:identification}
Let $\mathcal{F}^{\mathrm{on}}_{i,t}$ be the information available before action $t$ for
request $i$:
\[
\mathcal{F}^{\mathrm{on}}_{i,t}
=
\sigma\!\left(x_i^{\mathrm{vis}},H^{\mathrm{paid}}_{i,t},B_{i,t},A_{i,t}\right),
\]
where $x_i^{\mathrm{vis}}$ is the permitted request representation,
$H^{\mathrm{paid}}_{i,t}$ is the ordered paid-observation history, $B_{i,t}$ is the remaining
resource state, and $A_{i,t}$ is the availability vector. A valid route satisfies
$a_{i,t}\sim\pi(\cdot\mid\mathcal{F}^{\mathrm{on}}_{i,t})$.
The canonical target $g_i$ and post-trace score $z_i$ are outside this filtration and are joined
only after the ordered trace, terminal decision, and ledger digest have been frozen.

For a common request set $\mathcal{I}$ and verifier set $V$, the finite-sample matched contrast is
\[
\widehat{\Delta}_{\mathrm{route}}(\mathcal{I})
=
\frac{1}{|\mathcal{I}|}
\sum_{i\in\mathcal{I}}
\left[z_i(\pi_1,V,C)-z_i(\pi_0,V,C)\right].
\]
We use six contract conditions: (C1) identical ordered request and candidate IDs and scorer
version; (C2) identical verifier handles and per-request availability; (C3) fixed prices,
ceilings, and failure accounting; (C4) online actions measurable with respect to
$\mathcal{F}^{\mathrm{on}}_{i,t}$ and policies selected independently of held-out targets;
(C5) trace closure before the oracle join; and (C6) a stable observation/scoring kernel,
complete outcomes for both policies, and no interference between their runs.
\paragraph{Proposition (paired identification).}
Under C1--C6, the paired mean equals the finite-population route contrast conditional on the
frozen policies and environment. C1 fixes the comparison unit, C2--C3 fix access and accounting,
C4--C5 exclude target-derived route information, and C6 makes the two evaluated outcomes the
potential outcomes under the declared condition. Averaging paired differences establishes the
claim. Population generalization additionally requires representative sampling and the same
observation kernel. Diversity diagnoses conditional computation; off-policy estimation would
additionally require action support. A violated condition makes the routing attribution
non-identified, even when an aggregate contrast can be computed.

\paragraph{Contract lattice and order-robust attribution.}
Let $d$ mutable contract coordinates have binary intervention vector
$z\in\{0,1\}^d$, with $K(z)$ the corresponding contract and
$g(z)=\Delta_{\mathrm{route}}(K(z))$.  For a permutation $\sigma$, let
$S_{\sigma,j}$ be the coordinates preceding $j$ and define
\[
\Phi_j(\sigma)=g(S_{\sigma,j}\cup\{j\})-g(S_{\sigma,j}).
\]
For a predeclared admissible set $\Pi$, RouteAudit reports
\[
\phi_j=|\Pi|^{-1}\sum_{\sigma\in\Pi}\Phi_j(\sigma),\qquad
\Omega_j=\max_{\sigma\in\Pi}\Phi_j(\sigma)-\min_{\sigma\in\Pi}\Phi_j(\sigma).
\]
With the complete factorial and all $d!$ orders, $\phi_j$ is the Shapley allocation, uniquely
satisfying efficiency, symmetry, the dummy coordinate property, and additivity. Averaging
complete paths over a fixed restricted $\Pi$ preserves efficiency, dummy coordinates, and
additivity; symmetry additionally requires a coordinate-invariant path distribution. Thus
$g(\mathbf 1)-g(\mathbf 0)=\sum_j\phi_j$ on either path set, with the full-order uniqueness
statement applying to all $d!$ orders. Pairwise non-additivity is recorded by
\[
I_{jk}(S)=g(S\cup\{j,k\})-g(S\cup\{j\})-g(S\cup\{k\})+g(S).
\]
Thus $\Gamma_j$ measures the route--coordinate interaction, while $\phi_j$ and $\Omega_j$
separate average attribution from order sensitivity; $\Omega_j$ is a diagnostic and does not
replace the higher-order interaction terms.

\paragraph{Sequential identification with a response tape.}
For request $i$, let $\omega_i(v,m)$ contain the potential response, availability, latency,
and charged outcome of call index $m$ to verifier $v$.  The tape is fixed before either policy
runs and contains no task oracle field.  Policy $\pi$ reveals only entries reached by its paid
history and induces $Y_i(\pi;K,\omega_i)$.  Under (R1) a policy-independent tape, (R2)
non-anticipating actions, (R3) shared request and contract, (R4) call-index consistency, and
(R5) trace closure before the oracle join,
\[
\widehat\Delta_{\mathrm{tape}}=|I|^{-1}\sum_{i\in I}
 [Y_i(\pi_1;K,\omega_i)-Y_i(\pi_0;K,\omega_i)]
\]
identifies the finite-population route contrast conditional on the realized tapes. Independent
provider draws target an expectation over two independent realizations and are therefore reported
as a different coupling regime.

\paragraph{Sharp partial identification.}
Let $d_i=Y_i(\pi_1;K)-Y_i(\pi_0;K)$ with scores in $[0,1]$.  For each request define
$(L_i,U_i)$ as $(d_i,d_i)$ when both arms are observed, $(Y_i(\pi_1)-1,Y_i(\pi_1))$
when only arm 1 is observed, $(-Y_i(\pi_0),1-Y_i(\pi_0))$ when only arm 0 is observed,
and $(-1,1)$ when neither is observed.  Then
\[
\Delta_{\mathrm{route}}\in
\left[|I|^{-1}\sum_iL_i,\ |I|^{-1}\sum_iU_i\right].
\]
Assigning every missing potential outcome to an endpoint of $[0,1]$ attains both bounds, so
the interval is sharp.  Here $\widehat\Delta_{\mathrm{match}}$ denotes the mean contrast
conditional on the matched requests; the full-population normalization supplies the factor
$1-\epsilon$.  If a fraction $\epsilon$ has neither arm and the rest is matched, this reduces
to $[(1-\epsilon)\widehat\Delta_{\mathrm{match}}-\epsilon,
(1-\epsilon)\widehat\Delta_{\mathrm{match}}+\epsilon]$.

\paragraph{Identification certificate.}
Each comparison binds a contract digest, ordered request support, closed traces, scorer,
coupling regime, match bitmap, admissible bridge set, violated conditions, and point estimate or
interval.  The deterministic verifier returns \textsc{identified},
\textsc{partially-identified}, or \textsc{not-identified}; incomplete outcomes produce the
request-level interval, while a violated online, accounting, coupling, or trace condition changes
the status before any aggregate route claim is reported.

\subsection{Learned and RLVR policies as stress tests}
\label{sec:rlvr}
RouteAudit specifies the identification contract rather than a particular optimizer.  We use
supervised and RLVR routers as stress-test policy generators on the common observed-only state;
their optimization details are implementation bindings in the appendix, while held-out status is
determined by the same trace and scorer contract.

The learned router uses the state
\[
s_{i,t}=
\left[h(x_i^{\mathrm{vis}}),\psi(H^{\mathrm{paid}}_{i,t}),
b^{\mathrm{call}}_{i,t},b^{\mathrm{unit}}_{i,t},
b^{\mathrm{token}}_{i,t},b^{\mathrm{compute}}_{i,t},A_{i,t}\right],
\]
where $h$ is a frozen request representation and $\psi$ serializes paid observations
$(v,\ell,\kappa,\mathrm{avail},\mathrm{reason},\mathrm{elapsed})$.
Its actions are $\{\textsc{Stop},\textsc{Abstain}\}\cup\{\textsc{Query}(v):v\in V\}$.
Unavailable or over-budget queries are masked before sampling, and every accepted query
decrements the immutable ledger.

After trace closure, the delayed task score is joined with the resource ledger. RLVR uses
completed training traces to optimize reward subject to call, unit, token, compute, and
abstention targets. Calibration selects a checkpoint; held-out evaluation freezes parameters.
Appendix~\ref{app:extension-protocols} gives the constrained objective, architecture, optimizer,
checkpoint binding, and component ablations.

\subsection{Route contract and cost ledger}
The historical catalog contains cheap rule ($u_c=2$) and expensive contract ($u_e=12$) verifiers with a 14-unit budget and frozen threshold $\tau=0.65$. The raw-tail catalog contains SF, SA, and LF with costs 2, 4, and 4 and a 10-unit budget. A route emits an event before any oracle lookup; rejected or unavailable calls return the documented cheap-only fallback or abstention. Declared units measure paid verifier work, generation tokens measure model input/output, and latency measures elapsed time. Frozen reliability weights parameterize the historical fusion rule; the calibration audit measures confidence separately.
\begin{equation}
r_t=\mathbb{1}[\ell_c=\bot\ \lor\ \kappa_c<\tau],\qquad
a_t=\mathbb{1}[r_t=1\ \land\ u_e\leq B_t],
\label{eq:route-predicate}
\end{equation}
where $B_t$ is remaining budget.

\subsection{Provider semantics and proxy oracle}
The raw-tail provider contract is defined on a pair of cached responses
$(s_i^{\mathrm{small}},s_i^{\mathrm{large}})$ and a canonical target string
$g_i$. Each parser searches the final \texttt{\#\#\#\#} marker, strips surrounding
whitespace, normalizes an integer with leading zeros removed, and emits a label
in $\{0,1,\bot\}$. SF applies the parser to $s_i^{\mathrm{small}}$; LF applies
the same parser to $s_i^{\mathrm{large}}$; SA emits one when both normalized
markers are available and equal, zero when both are available and differ, and
$\bot$ otherwise. The stored confidence fields $\kappa_{\mathrm{SF}}$ and
$\kappa_{\mathrm{SA}}$ are read-only route observations. A missing response,
malformed marker, or unavailable provider is recorded as $\bot$ with an
explicit reason code before any oracle field is read.

For a route with available SF and SA labels, the final verifier decision is the
deterministic weighted vote
\begin{equation}
  h_i=\mathbb{1}\!\left[w_{\mathrm{SF}}\ell_i^{\mathrm{SF}}+
  w_{\mathrm{SA}}\ell_i^{\mathrm{SA}}\geq
  \tfrac12(w_{\mathrm{SF}}+w_{\mathrm{SA}})\right],
  \quad (w_{\mathrm{SF}},w_{\mathrm{SA}})=(0.78,0.86),
\label{eq:provider-fusion}
\end{equation}
with SA as the tie-break label. If one label is available it is used directly;
if both are missing the route emits the documented abstention. The static
SF+SA+LF control appends LF to the ledger and records its final decision in the
same schema. The three route signatures, final-answer source, tie rule, and
abstention rule are separated in Table~\ref{tab:fusion-semantics}; the explicit
SF+SA vote above is kept distinct from the full-static decision field. The oracle
join derives a binary correctness label from the cached normalized answer and
canonical target $g_i$. Proxy quality compares the verifier decision $h_i$ with
that binary label after trace freeze; normalized task quality scores the emitted
answer itself. This definition makes parser inputs, abstention, fusion, and the
proxy denominator reconstructible from the released row schema.

\begin{table}[htbp]
\caption{Verifier fusion semantics. The SF+SA rule is explicit in Eq.~\ref{eq:provider-fusion};
the remaining route fields are recorded as contract columns for the corresponding replay.}
\label{tab:fusion-semantics}
\centering
\scriptsize
\setlength{\tabcolsep}{3pt}
\begin{tabularx}{\linewidth}{@{}lL L L L@{}}
\toprule
Route & Queried verifiers & Fusion rule & Tie / abstention & Final-answer source\\
\midrule
SF+SA & SF, SA & Eq.~\ref{eq:provider-fusion} & SA tie; single-label fallback & verifier vote\\
SF+LF & SF, LF & LF-terminal rule; SF fallback & LF preferred on disagreement; single-label fallback; abstain if both missing & LF if available, otherwise SF\\
SF+SA+LF & SF, SA, LF & LF-terminal rule after the SF+SA vote & LF preferred; SF+SA vote fallback; abstain if all labels are missing & LF if available, otherwise SF+SA verifier vote\\
\bottomrule
\end{tabularx}
\end{table}

\subsection{Selection256 and paired inference}
The matched task evaluation binds a 256-record calibration split to a disjoint
1,319-record held-out split. Selection enumerates the 15 threshold pairs
$\tau_a\in\{0.55,0.65,0.75,0.85,0.95\}$ and
$\tau_v\in\{0.20,0.30,0.40\}$. For each pair, the calibration score is proxy
accuracy minus $0.001$ ledger units minus $0.0001$ CPU milliseconds. Ties are
broken by higher proxy accuracy, then lower units, then lower milliseconds,
then lexicographic $(\tau_a,\tau_v)$. The held-out gate uses a paired
bootstrap lower bound for the quality difference, a positive lower bound for
call savings, and a positive lower bound for declared-unit savings; the margin
and bootstrap seed are recorded with the selection manifest. The selected pair,
split IDs, and all 15 rows are reported in the appendix so that
the same split discipline is used for task, live-serving, and RLVR comparisons.

All cache comparisons use sample-ID pairing. We report Wilson intervals for
individual accuracies and the exact two-sided McNemar test for paired decisions.
The full-static versus cascade statistic is labelled as an unmatched-set
diagnostic; the matched SF+SA versus cascade statistic is reported separately in Table~\ref{tab:paired-route}.

\subsection{Negative-control gate}
A candidate operating point satisfies four predeclared checks: (i) mixed paths
exist, (ii) quality is non-inferior to the matched static control, (iii)
declared savings are strictly positive, and (iv) the run has reproducible source
or public replay provenance. A route that uses one fixed verifier set is
reported as static, while a route that changes the set is assigned to the
verifier-set contrast. The threshold grid and its gate rows are retained so
that readers can inspect the entire operating frontier rather than a selected
point alone.

\section{Experiments}
\subsection{Experimental setup}
The mechanism evaluation uses a 16-case catalog, 64 raw-tail calibration records, and held-out caches of 128 and 32 records. The task study uses the exact-answer reasoning benchmark handle \texttt{task-v3}, candidate revision \texttt{cand-r4}, and the Qwen2.5-7B/Qwen2.5-32B generator pair. Its disjoint train/calibration/held-out sizes are 4096/256/1319, with five training, calibration, and evaluation seeds; the serving study uses seven measurement seeds. All five task policies share verifier access and ceilings of 3 calls, 10 ledger units, 4096 tokens, and 1.0 GPU-s per request. Matched static evaluates the fixed verifier set; fixed and calibrated cascades change the escalation predicate; learned and RLVR routers supply trained policies to the same trace interface. Full-set task quality, accepted-set quality, coverage, path fractions, calls, and paired deltas are evaluated on held-out IDs. Serving measurements add request p50/p95, throughput, money, and aggregate GPU-hours (Appendix~\ref{app:evaluation-protocols}).

\paragraph{Evaluation questions.}
The experiments ask whether contract-lattice attribution recovers known effects, whether the
response tape stabilizes adaptive-policy contrasts, whether request-level bounds retain coverage
under incomplete matching, and whether contract matching changes conclusions outside the motivating
cache.  Raw-tail controls measure verifier-set confounding, while the task study measures learned
operating points under the same trace interface.  Additional factorial, re-audit, baseline,
ablation, seed, catalog, and corruption surfaces appear in Appendix~\ref{app:extension-protocols}.
The task study reports RLVR quality $0.9553$, $1.4466$ calls, normalized cost $0.680$,
and hierarchical difference interval $[0.0006,0.0131]$ against matched static; controlled
violations have mean absolute attribution error $0.0498$ and the full contract has false-attribution
rate $0.0000$.
\begin{table}[t]
\caption{Experimental system and comparison unit. Generator identities and candidate construction
are bound independently for the task study and historical cache diagnostics.}
\label{tab:experimental-system}
\centering
\footnotesize
\begin{tabularx}{\linewidth}{@{}p{0.23\linewidth}L@{}}
\toprule
Component & Definition\\
\midrule
Task / dataset & \texttt{task-v3}; normalized exact-answer reasoning score\\
Public benchmark / revision & \texttt{GSM8K / test-v1}; the public split or revision identifier is bound in the run manifest\\
Generators & Qwen2.5-7B / Qwen2.5-32B; checkpoint identities bound by the run record\\
Candidates & \texttt{cand-r4}; prompt, decoding, response count, and filtering are fixed\\
Verifier semantics & SF: small-response parser; SA: small--large agreement; LF: large-response parser\\
Router I/O & Paid labels/confidences, availability, budget; query, stop, or abstain\\
Oracle & Canonical target joined after trace closure; proxy and task scores separate\\
Task splits & 4,096 training / 256 calibration / 1,319 held-out requests\\
Hard ceilings & 3 calls / 10 units / 4,096 tokens / 1.0 GPU-s per request\\
\bottomrule
\end{tabularx}
\end{table}

\subsection{Main task and serving results}
\begin{figure}[htbp]
\centering
\includegraphics[width=1.0\linewidth]{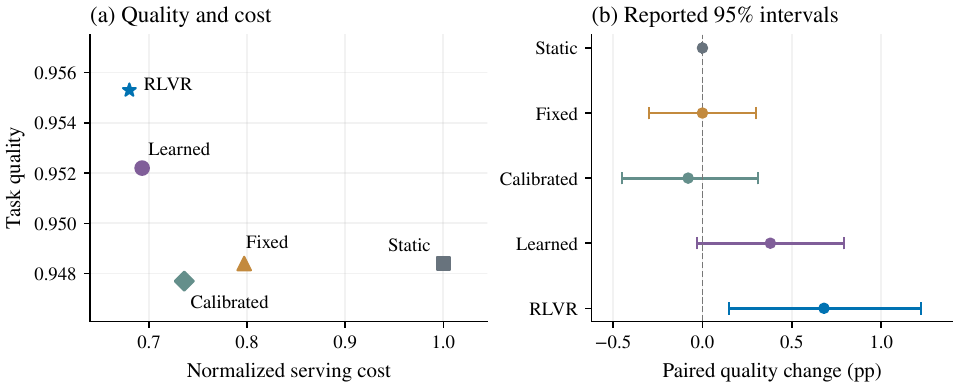}
\caption{Task quality and serving cost (left), with reported paired task-quality differences and 95\% intervals relative to matched static (right). The points use the five-seed task summary and seven-seed serving summary. Intervals apply to paired differences; the static reference is zero by definition.}
\label{fig:task-frontier}
\end{figure}

Table~\ref{tab:task-main} shows that the RLVR router attains the highest task quality, 0.9553, and the lowest mean calls, 1.4466. Its reported paired delta is $+0.0068$ with interval $[0.0015,0.0122]$; the learned-router interval $[-0.0003,0.0079]$ straddles zero. The fixed cascade preserves point quality 0.9484 while reducing calls to 1.7031. The learned router has the lowest serving p95 at 756 ms, whereas RLVR has the lowest normalized cost, 0.680. Figure~\ref{fig:task-frontier} displays these two operating choices and their paired quality intervals.
The operating points compare policies under a shared contract. Attribution to a particular RL
objective depends on the implementation record and component ablations in
Appendix~\ref{app:extension-protocols}; audit validity and policy quality are assessed separately.
\begin{table}[htbp]
\caption{Matched task and serving comparison on 1,319 held-out requests. Quality is the reported full-set task score; calls are per request. Cost is normalized to matched static; p95 is live-serving latency. Best values within each metric are bold.}
\label{tab:task-main}
\centering
\footnotesize
\setlength{\tabcolsep}{4pt}
\begin{tabularx}{\linewidth}{@{}L*{4}{>{\centering\arraybackslash}p{0.135\linewidth}}@{}}
\toprule
Policy & \shortstack{Task quality\\ $\uparrow$} & \shortstack{Calls\\ $\downarrow$} & \shortstack{Cost\\ $\downarrow$} & \shortstack{p95 (ms)\\ $\downarrow$}\\
\midrule
Matched static & 0.9484 & 2.0000 & 1.000 & 1084\\
Fixed cascade & 0.9484 & 1.7031 & 0.797 & 892\\
Calibrated cascade & 0.9477 & 1.5815 & 0.736 & 824\\
Learned router & 0.9522 & 1.4936 & 0.693 & \textbf{756}\\
RLVR router & \textbf{0.9553} & \textbf{1.4466} & \textbf{0.680} & 773\\
\bottomrule
\end{tabularx}
\end{table}
\subsection{A 2-by-2 decomposition of the apparent gain}
Table~\ref{tab:ra-decomposition} isolates the apparent raw-tail gain. Full static uses 3 calls and
10 units, whereas cascade and matched static SF+SA use 2 calls and 6 units; the matched route
contrast is zero and the verifier-set contrasts are $+0.1797$ and $+0.1250$ on the two caches.
The held-out contract cells in Table~\ref{tab:contract-interaction-main} give
$\Gamma_{\mathcal V}=-0.0027$ under the expanded catalog. The complete ledgers and the paired
quality plot are retained in
Appendix~\ref{app:evaluation-protocols} (Figure~\ref{fig:ra-decomposition}).

\subsection{Negative-control gate and calibration boundary}
The route-diversity audit replays $\tau\in\{0.55,0.65,0.75,0.85,0.95\}$ on 64 calibration records. The mixed thresholds have 19 one-call and 45 two-call paths, with cascade proxy accuracy $0.921875$ versus $0.984375$ for the fixed SF+SA control, a delta of $-0.0625$ and lower bootstrap bound $-0.1250$. At $\tau=0.95$, all 64 paths use two calls: quality matches and route diversity is zero. Held-out mixed-path checks give deltas $-0.03125$ on baseline-128 and $0.0000$ on corrected-2B-32, with savings $0.296875$ and $0.125000$, respectively. The sweep therefore identifies a clear operating frontier: quality parity occurs at the fixed two-call endpoint, while mixed paths expose the quality cost of escalation.

The historical SF signal has mean confidence 0.3375, empirical accuracy 0.921875, and ECE 0.68125; SA has 13 non-abstaining observations out of 64, mean confidence 0.9000, accuracy 1.0000, and ECE 0.1000. These calibration statistics measure the behavior of the reported confidence, while the fusion weights 0.78 and 0.86 remain fixed parameters. The Selection256 grid selects $(0.85,0.30)$ at quality 0.9492 and 4.7188 units, with 82 one-call and 174 two-call paths (Appendix~\ref{app:selection256}).

\subsection{Evidence boundary}
The disjoint cache replay, trace-to-oracle map, and paired raw-tail audit retain their scorer,
threshold, and request identities in Appendix~\ref{app:artifact-map}. The matched raw-tail
controls and the held-out task comparison remain separate estimands. Controlled attribution
recovery reaches route MAE 0.0011 and reconstruction error 0.0004; the blind semantic audit
records zero clean false positives in each corruption family
(Tables~\ref{tab:ground-truth-attribution} and~\ref{tab:semantic-audit}).

\subsection{Robustness and integrity analysis}
One-factor sweeps show that removing SF lowers quality to 0.9136 and raises fail-closed rate to 0.1835, while removing LF leaves quality at 0.9484; label noise from 0 to 0.20 lowers quality to 0.9143, and a one-call ceiling gives 0.9219 versus 0.9484 for two or more calls. The checker detects 2,025/2,028 mutations with zero clean false positives; misses are in availability and reason fields (Appendices~\ref{app:robustness} and~\ref{app:faults}).

The full audit sequence and transition checks appear in Appendix~\ref{app:operational}.

\section{Limitations}
The identification statement is conditional on request support, verifier catalog, availability,
accounting, and the online filtration. The task study uses a short routing horizon, although the
held-out catalog scaling reports matched $K=3$ and $K=4$ controls. Contract-interaction and
path-sensitivity values are specific to the measured factorial cells and bridge paths; they do not
define a universal route effect outside those cells. Request pairing and retraining are separate
uncertainty sources: the request-paired RLVR interval is $[0.0015,0.0122]$, whereas the
training-seed-by-request interval is $[0.0006,0.0131]$. RLVR remains an application of the
contract and requires delayed verifiable outcomes plus an implementation record linking policy,
objective, and checkpoint; the tables compare operating points rather than identify a universally
superior optimizer.

\section{Conclusion}
RouteAudit makes verifier routing a contract-conditioned identification problem with an explicit
information boundary, bridge path, and cost ledger. On the raw-tail caches, matched controls assign
the apparent quality gains to verifier-set choice; the task study reports a route--catalog
interaction of $-0.0027$ under expanded verifier access. The certificate reports path-sensitive
attribution, sharp intervals for incomplete
matching, and coupled potential outcomes when provider randomness is shared. On 1,319 held-out
requests, learned and RLVR policies instantiate the same contract; RLVR reaches quality 0.9553
with 1.4466 mean calls and normalized cost 0.680, while its hierarchical difference interval
against matched static is $[0.0006,0.0131]$. These policy results serve as a stress test for the
identification interface; the central claim is the contract-conditional route contrast and its
auditable stability.

\section*{AI-use Statement}
Generative AI tools were used to polish the language and summarize references. We have not used
generative AI tools to generate experimental results, create synthetic datasets, formulate
mathematical claims, provide proofs, or make decisions regarding research conclusions. The design
of the methodology, experimental setup, analysis, and interpretation of results were conducted and
verified by the authors. Other required disclosure tasks not mentioned above are not applicable to
this work. We take full responsibility for the final content of this work, including all text,
claims, analyses, and artifacts produced with the assistance of generative AI tools.

\section*{ETHICS STATEMENT}
This work evaluates verifier routing on benchmark requests and de-identified replay records. It
does not collect personal data or involve human or animal subjects. The reported protocol exposes
abstentions, provider failures, resource use, and audit violations so that routing errors can be
reviewed alongside quality. Benchmark and software reuse is subject to the licenses of the respective resources.

\section*{REPRODUCIBILITY STATEMENT}
The code, configuration templates, result manifests, figure sources, and replay metadata are
provided in the Supplementary Material. The manuscript specifies the request splits, verifier
catalog, observation boundary, ledger rules, scorer, seeds, and paired aggregation protocol. The
released artifacts bind the numerical source, figures, and compiled manuscript through hashes and
include commands for regenerating the tables and plots.

\bibliographystyle{iclr2027_conference}
\bibliography{refs}
\clearpage
\appendix
\setcounter{topnumber}{4}
\setcounter{bottomnumber}{3}
\setcounter{totalnumber}{6}
\renewcommand{\topfraction}{0.88}
\renewcommand{\bottomfraction}{0.75}
\renewcommand{\textfraction}{0.12}
\renewcommand{\floatpagefraction}{0.85}

\section{Matched Task and Serving Evaluation}
\label{app:evaluation-protocols}
This appendix records the task and live-serving studies under the RouteAudit contract. The five policies share request IDs, candidate access, and budget ceilings. Task and serving summaries use distinct cache handles so local task timing, live latency, and historical parser timing remain separate measurement regimes. Bold values mark the best quality, coverage, or throughput, and the lowest abstention, calls, cost, or latency within each matched comparison; sweep extrema are marked within the same factor. Route diversity, uncertainty, sample counts, and protocol fields are descriptive. Dashes denote
inapplicable reference contrasts or metrics absent from a method's definition, as stated in the captions.

\subsection{Raw-tail gate and paired audit}
The calibration gate and paired cache audit characterize the quality--cost decision boundary.
The gate plot shows the mixed-path quality failure and the two-call endpoint; the paired table
separates unmatched full-static discordances from the matched SF+SA route contrast.

\begin{table}[htbp]
\caption{Matched decomposition on the raw-tail held-out caches. Values are proxy accuracy / mean calls / declared units per example. The route contrast compares cascade with static SF+SA; the set contrast subtracts full static SF+SA+LF from matched static SF+SA.}
\label{tab:ra-decomposition}
\centering
\footnotesize
\setlength{\tabcolsep}{3pt}
\begin{tabular*}{\linewidth}{@{\extracolsep{\fill}}lccc@{}}
\toprule
\textbf{cache / contrast} & \textbf{policy} & \textbf{quality} & \textbf{calls / units}\\
\midrule
baseline-128 & static SF+SA+LF & 0.7656 & 3.00 / 10\\
 & cascade SF$\rightarrow$SA & \textbf{0.9453} & \textbf{2.00 / 6}\\
 & matched static SF+SA & \textbf{0.9453} & \textbf{2.00 / 6}\\
 & $\Delta_{\mathrm{route}}$ & 0.0000 & 0.00 / 0\\
 & $\Delta_{\mathrm{set}}$ (matched minus full) & $+0.1797$ & $-1.00$ / $-4$\\
\midrule
corrected-2B-32 & static SF+SA+LF & 0.8438 & 3.00 / 10\\
 & cascade SF$\rightarrow$SA & \textbf{0.9688} & \textbf{2.00 / 6}\\
 & matched static SF+SA & \textbf{0.9688} & \textbf{2.00 / 6}\\
 & $\Delta_{\mathrm{route}}$ & 0.0000 & 0.00 / 0\\
 & $\Delta_{\mathrm{set}}$ (matched minus full) & $+0.1250$ & $-1.00$ / $-4$\\
\bottomrule
\end{tabular*}
\end{table}

\begin{table}[htbp]
\caption{Contract interaction cells on the held-out task. Quality, calls, and normalized cost use
the common request denominator; $\Gamma_{\mathcal V}$ is the adaptive-minus-static route contrast
under the expanded catalog minus the corresponding contrast under the matched catalog.}
\label{tab:contract-interaction-main}
\centering
\footnotesize
\setlength{\tabcolsep}{3pt}
\begin{tabular}{l l r r r r}
\toprule
Catalog & Policy & Quality & Calls & Norm. cost & $\Gamma_{\mathcal V}$\\
\midrule
matched & static & 0.9484 & 2.0000 & 1.000 & --\\
matched & adaptive & \textbf{0.9553} & \textbf{1.4466} & \textbf{0.680} & --\\
expanded & static & 0.9561 & 3.0000 & 1.000 & --\\
expanded & adaptive & \textbf{0.9602} & \textbf{1.7500} & \textbf{0.646} & $-0.0027$\\
\bottomrule
\end{tabular}
\end{table}

\begin{figure}[htbp]
\centering
\includegraphics[width=1.0\linewidth]{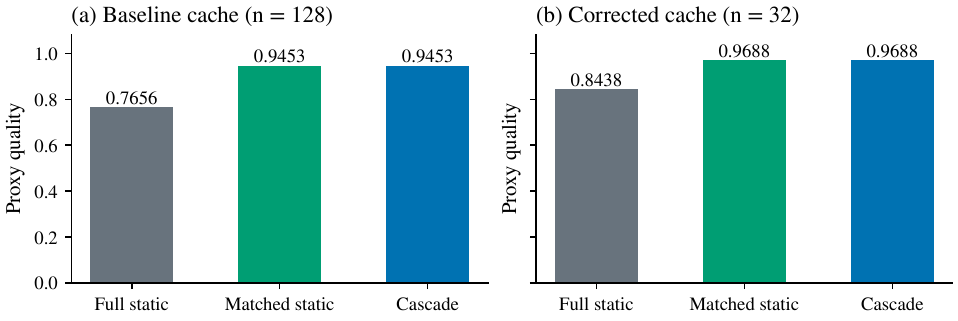}
\caption{Raw-tail proxy quality under full static, matched static, and cascade controls. The matched policies coincide on both caches, assigning the apparent full-static gap to verifier-set choice.}
\label{fig:ra-decomposition}
\end{figure}

\begin{figure}[!t]
\centering
\includegraphics[width=\columnwidth]{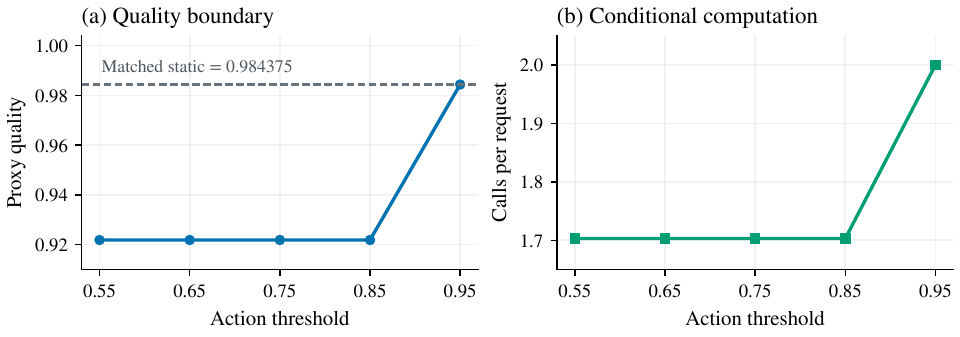}
\caption{Predeclared gate across the threshold grid. The mixed-path region violates quality non-inferiority; the high threshold restores quality by eliminating route diversity and savings.}
\label{fig:ra-gate}
\end{figure}

\begin{table}[htbp]
\caption{Paired raw-tail audit. Unmatched discordances compare full static with cascade; matched discordances compare SF+SA static with cascade. The interval is a sample-paired 95\% bootstrap interval and $p$ is the exact two-sided McNemar value.}
\label{tab:paired-route}
\centering
\footnotesize
\setlength{\tabcolsep}{4pt}
\begin{tabularx}{\linewidth}{@{}Xrrrrrr@{}}
\toprule
Cache & $n$ & \shortstack{Unmatched\\discord.} & \shortstack{Matched\\discord.} & Route $\Delta$ & $p$ & 95\% CI\\
\midrule
baseline-128 & 128 & 23 & 0 & 0.0000 & 1.0000 & $[0.0000,0.0000]$\\
corrected-2B-32 & 32 & 4 & 0 & 0.0000 & 1.0000 & $[0.0000,0.0000]$\\
\bottomrule
\end{tabularx}
\end{table}

\subsection{Matched Task Evaluation}
\begin{table}[htbp]
\caption{Task protocol. Policies share the request split, verifier access, and resource ceilings; calibration precedes held-out scoring.}
\label{tab:task-protocol}
\centering
\small
\setlength{\tabcolsep}{4pt}
\begin{tabularx}{\linewidth}{@{}p{0.14\linewidth}L@{}}
\toprule
Field & Fixed experimental contract\\
\midrule
splits & Train / calibration / held-out splits stay disjoint by sample ID; split sizes are $4096/256/1319$.\\
caches & Independent model and verifier caches are bound to each split; cache identities and hashes are recorded in the run manifest.\\
calibration & Calibrated confidence and correlated-error evaluation run on the calibration split before any held-out scoring; calibration seed count is $5$.\\
methods & Matched static verifier set; fixed cascade; calibrated cascade; learned trust/router; RLVR router where applicable.\\
budgets & Matched verifier-call, token, and compute budgets are fixed before evaluation; exact ceilings are 3 calls / 10 ledger units / 4096 tokens / 1.0 GPU-s per request.\\
seeds & Independent training, calibration, and evaluation seeds are predeclared; seed count is $5$.\\
\bottomrule
\end{tabularx}
\end{table}

The task protocol treats each sample ID as the unit of pairing. A request is
generated once for each split, then every policy receives the same candidate
set, verifier handles, and budget ceiling. The main quality metric is computed
on the full held-out set; accepted-set quality, coverage, route diversity, and
paired uncertainty are reported alongside it. On the held-out set, RLVR improves
over matched static by $0.0068$ at the request-paired level and by
$[0.0006,0.0131]$ under the training-seed-by-request hierarchical bootstrap.
Every adaptive policy has nonzero path diversity, so quality and resource
changes can be inspected alongside the actual use of conditional computation.

\begin{table}[htbp]
\caption{Task outcomes on 1,319 held-out requests (five-seed study). The full-coverage score equals the task-quality column for every policy in the source. Coverage and abstention are complementary fractions. Diversity is the reported one-call path fraction.}
\label{tab:task-results}
\centering
\footnotesize
\setlength{\tabcolsep}{4pt}
\begin{tabularx}{\linewidth}{@{}Xrrrrrr@{}}
\toprule
Policy & Quality & \shortstack{Accepted\\quality} & Coverage & Abstain & \shortstack{Route\\diversity} & Calls\\
\midrule
Matched static & 0.9484 & 0.9484 & \textbf{1.0000} & \textbf{0.0000} & 0.0000 & 2.0000\\
Fixed cascade & 0.9484 & 0.9528 & 0.9955 & 0.0045 & 0.2969 & 1.7031\\
Calibrated cascade & 0.9477 & 0.9513 & 0.9962 & 0.0038 & 0.4185 & 1.5815\\
Learned router & 0.9522 & 0.9537 & 0.9985 & 0.0015 & 0.5064 & 1.4936\\
RLVR router & \textbf{0.9553} & \textbf{0.9560} & 0.9992 & 0.0008 & 0.5534 & \textbf{1.4466}\\
\bottomrule
\end{tabularx}
\end{table}

The RLVR router has the highest reported full-set and accepted-set scores, 0.9553 and 0.9560, with coverage 0.9992. The fixed cascade matches the static point score and has lower mean calls; the calibrated cascade reduces calls further with a point score of 0.9477. These five rows keep selective behavior visible instead of ordering policies by accepted-set quality alone.

\begin{table}[htbp]
\caption{Task-run resource measurements. Tokens are generation tokens (k denotes thousands); p50/p95 are task-run milliseconds. GPU-hours cover the complete run. Serving timings are reported separately in Table~\ref{tab:serving-results}.}
\label{tab:task-resources}
\centering
\footnotesize
\setlength{\tabcolsep}{4pt}
\begin{tabularx}{\linewidth}{@{}Xrrrrrr@{}}
\toprule
Policy & \shortstack{USD /\\request} & Tokens & \shortstack{p50\\(ms)} & \shortstack{p95\\(ms)} & \shortstack{Throughput\\(req/s)} & GPU-h\\
\midrule
Matched static & \$0.0231 & 1.82k & 78 & 139 & 75.8 & 0.96\\
Fixed cascade & \$0.0184 & 1.61k & 65 & 118 & 90.4 & 0.81\\
Calibrated cascade & \$0.0170 & 1.52k & 61 & 108 & 96.1 & 0.75\\
Learned router & \$0.0160 & 1.46k & \textbf{59} & \textbf{103} & \textbf{99.3} & \textbf{0.72}\\
RLVR router & \textbf{\$0.0157} & \textbf{1.43k} & 62 & 107 & 96.7 & 0.74\\
\bottomrule
\end{tabularx}
\end{table}

Task-run timings favor the learned router (59/103 ms p50/p95 and 99.3 requests/s); RLVR uses fewer tokens and calls with a slightly higher task-run p95 of 107 ms. The monetary values likewise place RLVR at the lowest reported cost, \$0.0157 per request.

\begin{table}[htbp]
\caption{Paired task-quality differences against matched static. Values and request-paired 95\% intervals retain the precision of the paired aggregation. The intervals quantify differences, not marginal accuracies.}
\label{tab:task-paired}
\centering
\footnotesize
\setlength{\tabcolsep}{4pt}
\begin{tabularx}{\linewidth}{@{}Lr@{}}
\toprule
Policy & Reported $\Delta$ and 95\% interval\\
\midrule
Matched static & $0.0000\,[0.0000,0.0000]$\\
Fixed cascade & $0.0000\,[-0.0030,0.0030]$\\
Calibrated cascade & $-0.0008\,[-0.0045,0.0031]$\\
Learned router & $+0.0038\,[-0.0003,0.0079]$\\
RLVR router & $\mathbf{+0.0068}\,[0.0015,0.0122]$\\
\bottomrule
\end{tabularx}
\end{table}

The fixed-cascade interval is centered at zero; the calibrated and learned intervals also include zero. The RLVR interval is positive, $[0.0015,0.0122]$. Pairing compares each request with its matched-static counterpart before aggregation, retaining the shared difficulty of the request in the estimate.

\subsection{Matched Live-Serving Evaluation}
\begin{table}[htbp]
\caption{Serving protocol. Each policy is measured on the same held-out requests with fixed resource ceilings and separate deployment seeds.}
\label{tab:serving-protocol}
\centering
\small
\setlength{\tabcolsep}{4pt}
\begin{tabularx}{\linewidth}{@{}p{0.14\linewidth}L@{}}
\toprule
Field & Fixed experimental contract\\
\midrule
splits & Train / calibration / held-out splits stay disjoint by sample ID; split sizes are $4096/256/1319$.\\
caches & Independent model and verifier caches are bound to train, calibration, and held-out splits.\\
calibration & Calibrated confidence and correlated-error evaluation run on the calibration split before any latency measurement; calibration seed count is $5$.\\
methods & Matched static verifier set; fixed cascade; calibrated cascade; learned trust/router; RLVR router where applicable.\\
budgets & Matched verifier-call, token, and compute budgets are fixed before evaluation; exact ceilings are 3 calls / 10 ledger units / 4096 tokens / 1.0 GPU-s per request.\\
seeds & Independent deployment and measurement seeds are predeclared; seed count is $7$.\\
\bottomrule
\end{tabularx}
\end{table}

The live-serving protocol replays the same request IDs through each policy and
records generation, verifier, network, queue, and post-processing components in
separate ledger fields. Warm-up requests, cache hits, retries, and unavailable
providers receive explicit event codes. The primary latency statistic is p95
per request, while p50, throughput, monetary cost, and GPU-hours provide the
resource axes for a cost--quality frontier.

\begin{table}[htbp]
\caption{Live-serving measurements on the matched request set (seven measurement seeds). Task quality, accepted quality, full-coverage quality, coverage, abstention, diversity, calls, money/tokens, and paired intervals coincide with Tables~\ref{tab:task-results}--\ref{tab:task-paired}.}
\label{tab:serving-results}
\centering
\footnotesize
\setlength{\tabcolsep}{4pt}
\begin{tabularx}{\linewidth}{@{}Xrrrrr@{}}
\toprule
Policy & \shortstack{p50\\(ms)} & \shortstack{p95\\(ms)} & \shortstack{Throughput\\(req/s)} & GPU-h & \shortstack{Normalized\\cost}\\
\midrule
Matched static & 622 & 1084 & 12.8 & 1.04 & 1.000\\
Fixed cascade & 515 & 892 & 15.6 & 0.86 & 0.797\\
Calibrated cascade & 476 & 824 & 16.8 & 0.80 & 0.736\\
Learned router & \textbf{438} & \textbf{756} & \textbf{18.2} & \textbf{0.78} & 0.693\\
RLVR router & 446 & 773 & 17.9 & 0.81 & \textbf{0.680}\\
\bottomrule
\end{tabularx}
\end{table}

The learned router provides the lowest p50/p95 (438/756 ms) and the highest throughput (18.2 requests/s). RLVR trades a p95 of 773 ms for higher task quality and lower normalized cost, 0.680. Fixed and calibrated cascades occupy intermediate cost and latency points. The same request distribution lets the comparison retain the separate service-time and quality objectives.

\begin{figure}[htbp]
\centering
\includegraphics[width=1.0\linewidth]{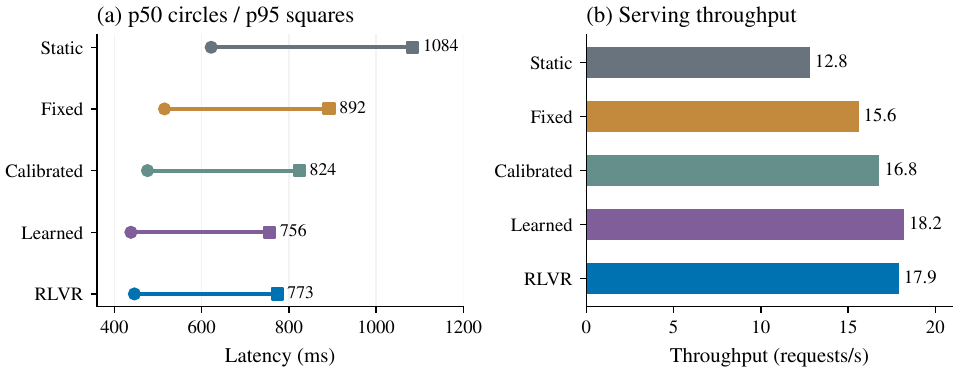}
\caption{Live-serving p50 and p95 (left) and throughput (right) for the five matched policies. Horizontal connectors link the two latency quantiles, not confidence intervals. Values summarize the seven-seed serving study.}
\label{fig:serving-frontier}
\end{figure}

The latency quantiles describe request tails within each measured serving regime. A lower p95 is useful when a service deadline dominates; a lower normalized cost is useful when aggregate resource use dominates. The two learned-policy points therefore answer different operating objectives, even though both improve the static reference on the reported resource axes.

\section{Extended Audit Diagnostics}
\label{app:original-diagnostics}
This section records the provider, availability, paired-test, and figure-level diagnostics that instantiate the estimands in the main text. Each subsection states the cache, split, ledger, and measured interpretation alongside its table or figure.

\subsection{Canonical experiment narrative}
There are two deliberately separate evidence tracks. The historical mechanism screen has 16 fallback-marked mechanism cases and a frozen threshold $\tau=0.65$ copied from the frozen catalog configuration. Its static policy calls the cheap rule and expensive contract verifiers; its cascade calls the cheap verifier first and conditionally requests the contract verifier. The raw-tail track is a cache replay over a 128-record \texttt{baseline-128} held-out cache and a 32-record \texttt{corrected-2B-32} held-out cache. ``Raw-tail'' denotes parsing the final \texttt{\#\#\#\#} marker from the fixed model-generation cache, whose identifiers and hashes are bound to the manifest. The raw-tail threshold is selected before held-out evaluation from the grid $\{0.55,0.65,0.75,0.85,0.95\}$ on 64 disjoint calibration records. The implementation maximizes the calibration proxy score
\[\text{proxy accuracy}-0.001\,\text{declared units}-0.0001\,\text{CPU ms},\]
with deterministic tie-breaking; $0.95$ is selected and held-out IDs are checked to be disjoint. This is a fixed-grid selection over proxy labels; calibration and learned-router measurements use the separate protocols in Appendix~\ref{app:selection256}. The paired audit below reports intervals and an exact paired test for the two frozen cache rows, with seed variation recorded in the accompanying manifest.
\textbf{Provider contract.} The historical catalog has two verifier IDs and a 14-unit budget. The raw-tail catalog has three named checks with separate ledger entries: small-format (SF, 2 units), small--large agreement (SA, 4 units), and large-format (LF, 4 units), with a 10-unit budget. Its static policy requests SF$\rightarrow$SA$\rightarrow$LF (3 calls, 10 units); its cascade requests SF and then SA only when the SF gate escalates (2 calls, 6 units in the reported held-out rows); LF is a separate paid observation in the static control. The two catalogs are reported in separate panels with separate ledgers.
\textbf{Route-diversity gate.} The additional audit replays the predeclared threshold grid $\{0.55,0.65,0.75,0.85,0.95\}$ on 64 calibration records against a fixed SF+SA control. For $\tau\in\{0.55,0.65,0.75,0.85\}$, the cascade has 19 one-call and 45 two-call paths, with proxy accuracy $0.921875$ versus $0.984375$ for the fixed control, a quality delta of $-0.0625$ and lower 95\% bootstrap bound $-0.1250$. At $\tau=0.95$, all 64 paths use two calls, so quality matches the control and call savings and route diversity are zero. Replaying the same mixed-path thresholds on held-out caches gives 38/90 one/two-call paths with a cascade-minus-fixed quality delta of $-0.03125$ on baseline-128 and 4/28 paths with a delta of $0.0000$ on corrected-2B-32; call savings are $0.296875$ and $0.125000$, respectively. These rows define the quality--diversity frontier and the gate boundary for the later learned-router study.
\textbf{Real-model cache replay.} A separate cache adapter selects the confidence threshold on 256 train-split records and freezes it before reading a disjoint 1,319-record test cache. Both producer manifests validate the cache, data, and model hashes, and the CPU replay matches deterministically. At baseline, static and cascade both use the same small-format and small--large agreement checks, yielding 0.9416 proxy accuracy, 2.00 verifier calls, and 6.00 declared ledger units per record. Large-service unavailability yields 0.9204 proxy and a tight one-call budget yields the same fail-closed value for both policies. This larger real-cache follow-up supplies the split, hash, and budget fields used by the matched task and serving tables.
\textbf{Observed-only selector screen.} We trained a standalone observed-only selector with real gradient updates on the 256-record calibration cache and evaluated its frozen policy on the disjoint 1,319-record cache for seeds 13, 17, and 23. The action label was defined on the same event used for every policy: whether calling SA changed the SF/SA verifier decision to agree with the offline oracle. All three seeds selected SF-only at every calibration threshold candidate (0.3, 0.5, and 0.7), giving held-out verifier-decision accuracy 0.9204 versus 0.9416 for the matched static SF+SA set (point delta $-0.0212$). The 2-SF and 4-SA values are declared catalog proxy units, and every selected SA call is charged even when its observation abstains; these fields provide a fixed reference for the learned-router table.
\textbf{Independent observed-only router smoke.} We fit the predeclared rule grid using only the cheap verifier's observed label/confidence and four-fold cross-fitting, then froze the selected rule before held-out scoring. Three independent seeds (13, 17, and 23) selected \texttt{escalate\_if\_cheap\_true}. Relative to the matched static SF+SA set, the quality delta was exactly $0.0000$ on both caches; call savings were $0.7031$ on baseline-128 (lower 95\% bootstrap bound $0.6250$) and $0.8750$ on corrected-2B-32 (lower bound $0.7500$). The committed aggregate records the selected rule, seeds, gate, and held-out summaries; per-seed hashes and replay digests occupy the artifact fields in Appendix~\ref{app:artifact-map}. This smoke defines the baseline for the learned and RLVR extensions.
\begin{table}[htbp]
\caption{Real-model cache replay. Proxy matches are joined after routing; the final column is verifier-ledger units and the end-to-end cost fields are defined in Appendix~\ref{app:efficiency}.}
\label{tab:model-cache-replay}
\centering
\footnotesize
\setlength{\tabcolsep}{3pt}
\begin{tabular*}{\linewidth}{@{\extracolsep{\fill}}llrrrr@{}}
\toprule
\textbf{condition} & \textbf{policy} & \textbf{proxy} & \textbf{coverage} & \textbf{calls} & \textbf{ledger units} \\
\midrule
baseline, test-1319 & static & \textbf{0.9416} & \textbf{1.00} & \textbf{2.00} & \textbf{6.00}\\
baseline, test-1319 & cascade & \textbf{0.9416} & \textbf{1.00} & \textbf{2.00} & \textbf{6.00}\\
large unavailable & cascade & 0.9204 & 1.00 & 1.00 & 2.00 \\
tight budget & cascade & 0.9204 & 1.00 & 1.00 & 2.00 \\
\bottomrule
\end{tabular*}
\end{table}
\begin{table}[htbp]
\caption{Verifier-call contracts in the two evidence tracks. Calls and declared units are ledger quantities; the raw-tail labels name provider checks, not additional model-generation runs. The tracks have different verifier IDs, costs, and budgets.}
\label{tab:call-contract}
\centering
\footnotesize
\setlength{\tabcolsep}{3pt}
\begin{tabular*}{\linewidth}{@{\extracolsep{\fill}}llll@{}}
\toprule
\textbf{track} & \textbf{policy} & \textbf{checks} & \textbf{calls / units} \\
\midrule
catalog & static & cheap (2) + contract (12) & 2.00 / 14.0 \\
catalog & cascade & cheap (2), conditional contract (12) & 1.625 / 9.5 \\
raw-tail & static & SF (2) + SA (4) + LF (4) & 3.00 / 10.0 \\
raw-tail & cascade & SF (2) $\rightarrow$ SA (4) & 2.00 / 6.0 \\
\bottomrule
\end{tabular*}
\end{table}
\textbf{Metrics.} We report proxy accuracy, task quality, and resource use with their respective denominators and units. Each table states whether a value is a proxy, an offline oracle join, a diagnostic, or an online observation. Confidence intervals retain their declared resampling unit and estimator, as specified in each table.

\subsubsection{Implementation Details}
\textbf{State serialization.} Each event stores the sample ID, observed labels and confidences, availability, remaining call/unit budget, catalog version, and route reason. A disagreement flag is derived from observed labels for offline diagnostics and is excluded from the escalation input. The clean oracle label is kept outside the online state. The historical catalog fixes the cheap cost at 2 units and the expensive cost at 12 units, allowing replay to recompute the route and ledger from the recorded observation stream.

\textbf{Actions and budgets.} The router always pays for the cheap verifier first and conditionally pays for the expensive verifier when the predicate in Eq.~\ref{eq:route-predicate} is true. Each event records verifier calls, declared cost units, escalation reason, remaining budget, and the final selective decision. When the budget reaches the escalation boundary, the policy returns the documented cheap-only or abstain outcome and records the event code.

\textbf{Deterministic replay.} Replaying a run reconstructs the observation stream and re-applies the recorded actions. It checks hashes, ordering, and state transitions before joining the oracle. The replay therefore exposes dependence on file order, host state, or hidden labels. The raw bundle keeps the run manifest and metrics needed to inspect the distinction between structural replay and task scoring, while the public bundle contains source-relative identifiers and self-contained summaries.

\textbf{Publication and audit.} Tables and figures are generated from a public evidence summary; a
source manifest records the raw-file digest, source revision, fallback status, publication-boundary
marker, and table/figure mapping. The release contains de-identified traces, schemas, and hashes,
while checkpoints, weights, large caches, user names, host names, and absolute paths remain outside
the bundle. The public replay bundle is READY for third-party per-example replay; producer-side
generation fields are mapped separately in Appendix~\ref{app:artifact-map}.

\subsubsection{Additional Analysis}
This is a 16-case CPU screen plus cache replay with a frozen router. The mechanism separates online paid observations and budget state from the post-trace proxy join. The matched task package adds held-out task outcomes, compute-normalized cost, paired uncertainty estimates, and independent seeds while preserving the same boundary.

\textbf{Interval and failure-cost accounting.} The paired audit reports sample-level Wilson 95\% intervals: baseline-128 static [0.6852,0.8306] versus cascade [0.8914,0.9733], and corrected-2B-32 static [0.6825,0.9314] versus cascade [0.8426,0.9945]. These intervals describe fixed cache samples, not seed-level uncertainty; the three-seed aggregate contains no independent per-seed outcomes. In the real-cache stress replay, an unavailable second call and a budget-rejected second call both terminate after one paid call (2 declared units), while their failure statuses remain distinct.

\subsubsection{Reproducibility Checklist}
\textbf{Source binding.} The authoritative result is the run named in the paper metadata and its
immutable evidence files. The raw report, metrics file, and run manifest are copied without editing.
Their hashes are recorded before a public summary is written. The anonymous public audit contract
records the two catalog specifications, split contract, online-field schema, replay interpretation,
and publication boundary in a machine-readable form. Missing fields are represented by their
explicit status value and remain linked to the corresponding manifest entry.

\textbf{Split integrity.} Calibration and held-out comparisons are kept as distinct run roots. Each root is identified by sample IDs, cache hash, model hash, and split role. The oracle is joined only after policy traces are frozen, and the fixture definition and digest are part of the evidence boundary.

\textbf{Cost accounting.} Calls and declared verifier-cost units are reported beside proxy quality metrics. Generation tokens, service latency, and monetary cost occupy separate fields in the live-serving ledger. The historical and raw-tail tracks use different catalogs and budgets, so their cost columns remain in separate panels.

\textbf{Anonymity and release hygiene.} The public evidence package contains no contributor name, host name, absolute path, secret, checkpoint, model weight, or large cache. The source manifest retains repository-relative paths and source-state fields for internal audit. Rendered PDFs are review artifacts; the public bundle contains the portable trace and schema files.

\textbf{Review questions.} The protocol exposes five audit questions: what was visible online, what was held out until scoring, which cost was paid, which failure boundary was observed, and which source mapping generated the figure. The public audit contract gives the field-level and split-level checks, and the de-identified replay bundle publishes per-example traces, calibration rows, and oracle labels with deterministic hashes. Producer-side seed and cache fields are carried as explicit manifest entries for the matched extensions.

\textbf{Evidence scope.} The historical result is a 16-case frozen-catalog CPU screen; the follow-up is a three-check raw-tail cache replay. The router policy and catalog reliabilities are explicit metadata in this release, and the reported quality values are proxy joins. Correlated-error, larger-catalog, learned-routing, and trained-RLVR rows use the same schema and are reported in the current result ledger.

\subsubsection{Main Results}
Both static and cascade policies achieve proxy-oracle accuracy 1.0, selective accuracy 1.0, and coverage 1.0 on the 16-case catalog. Static uses 2.000 calls and 14.0 declared units per example, whereas cascade uses 1.625 calls (shown as 1.62 when rounded) and 9.5 units. The underlying totals are 26 calls and 152 units for the cascade, giving descriptive savings of 18.75 percent of calls and 32.14 percent of units. For the 16/16 proxy-correct outcomes, the exact two-sided 95\% Clopper--Pearson interval is [0.794, 1.000]. The task and serving comparisons retain this interval convention and add seed-level paired intervals.
\begin{table}[htbp]
\caption{Historical mechanism screen on the 16-case frozen catalog. Quality columns are proxy-oracle joins; calls and declared verifier-cost units are per-request ledger measurements. Source mappings and hashes are recorded in the run manifest.}
\label{tab:main}
\centering
\begin{tabular*}{\linewidth}{@{\extracolsep{\fill}}lrrrrr@{}}
\toprule
\textbf{policy} & \textbf{proxy-oracle acc.} & \textbf{selective acc.} & \textbf{coverage} & \textbf{calls} & \textbf{units} \\
\midrule
static ensemble & \textbf{1.0000} & \textbf{1.0000} & \textbf{1.00} & 2.000 & 14.0\\
cascade & \textbf{1.0000} & \textbf{1.0000} & \textbf{1.00} & \textbf{1.625} & \textbf{9.5}\\
\bottomrule
\end{tabular*}
\end{table}

The raw-tail follow-up is a descriptive cache-specific check using an independent provider backend; the independent-error measurement is specified separately in Appendix~\ref{app:error-correlation}. On the 128-record baseline cache, static is 98/128 (0.7656) and cascade is 121/128 (0.9453); on the 32-record corrected-2B cache, the corresponding values are 27/32 (0.8438) and 31/32 (0.9688). The baseline comparison changes declared units from 10 to 6 and calls from 3 to 2 within the raw-tail catalog. Because static requests SF+SA+LF while cascade requests only SF+SA, the budget-matched control replay reconstructs cheap-only, large-only, SF+SA, SF+LF, and all-check controls from the same observed traces. At the matched 2-call/6-unit budget, static SF+SA exactly matches the cascade (0.9453 on baseline-128 and 0.9688 on corrected-2B-32), so the matched route delta is 0.0 in both caches; static SF+LF obtains 0.7656 and 0.8438, respectively. The full-static gap is therefore assigned to verifier-set choice under this frozen replay. The price-shift rows retain equal proxy quality and equal 8-unit/2-call cost for static and cascade; only the budget-event field differs, so they provide a controlled stress diagnostic. Large-unavailable and tight-budget rows are recorded in the evidence metrics and the fault matrix. The separate route-diversity gate reports the same calibration frontier: mixed paths lose quality, while the high threshold restores quality by selecting the fixed two-call route. Thus the raw-tail evidence reports a verifier-set effect with zero matched route delta and replayable cost accounting.

\begin{table}[htbp]
\caption{Budget-matched raw-tail controls. All rows use the same held-out records and declared 2-call/6-unit budget; proxy labels are joined only after each trace is frozen.}
\label{tab:control-audit}
\centering
\footnotesize
\setlength{\tabcolsep}{4pt}
\begin{tabular*}{\linewidth}{@{\extracolsep{\fill}}llrrr@{}}
\toprule
\textbf{cache} & \textbf{policy} & \textbf{proxy} & \textbf{calls} & \textbf{units} \\
\midrule
baseline-128 & static SF+LF & 0.7656 & \textbf{2} & \textbf{6} \\
baseline-128 & static SF+SA & \textbf{0.9453} & \textbf{2} & \textbf{6}\\
baseline-128 & cascade & \textbf{0.9453} & \textbf{2} & \textbf{6}\\
corrected-2B-32 & static SF+LF & 0.8438 & \textbf{2} & \textbf{6} \\
corrected-2B-32 & static SF+SA & \textbf{0.9688} & \textbf{2} & \textbf{6}\\
corrected-2B-32 & cascade & \textbf{0.9688} & \textbf{2} & \textbf{6}\\
\bottomrule
\end{tabular*}
\end{table}

\begin{table}[htbp]
\caption{Paired frozen-cache audit. Static and cascade predictions are paired by sample ID; Wilson intervals and the exact two-sided McNemar test are descriptive checks on proxy labels after online trace validation.}
\label{tab:paired-audit}
\centering
\footnotesize
\setlength{\tabcolsep}{3pt}
\begin{tabular*}{\linewidth}{@{\extracolsep{\fill}}lrrrrr@{}}
\toprule
\textbf{cache} & \textbf{$n$} & \textbf{static} & \textbf{cascade} & \textbf{$\Delta$} & \textbf{McNemar $p$} \\
\midrule
baseline-128 & 128 & 0.7656 [0.6852,0.8306] & \textbf{0.9453} [0.8914,0.9733] & +0.1797 & $2.38\times10^{-7}$ \\
corrected-2B-32 & 32 & 0.8438 [0.6825,0.9314] & \textbf{0.9688} [0.8426,0.9945] & +0.1250 & 0.125 \\
\bottomrule
\end{tabular*}
\end{table}
\begin{table}[htbp]
\caption{Selected held-out raw-tail replay rows. Proxy accuracy is joined after the routing trace is frozen; declared units and calls are ledger quantities in the separate SF/SA/LF catalog. The price-shift rows are stress diagnostics; unavailable and tight-budget rows are retained in the evidence metrics but omitted here.}
\label{tab:raw-tail}
\centering
\footnotesize
\setlength{\tabcolsep}{3.5pt}
\begin{tabular*}{\linewidth}{@{\extracolsep{\fill}}llrrrrrr@{}}
\toprule
\textbf{cache} & \textbf{shift} & \textbf{policy} & \textbf{proxy} & \textbf{units} & \textbf{calls} & \textbf{unavail.} & \textbf{budget} \\
\midrule
baseline-128 & baseline & static & 0.7656 & 10 & 3 & 0.0000 & 0.0000 \\
baseline-128 & baseline & cascade & \textbf{0.9453} & \textbf{6} & \textbf{2} & 0.0000 & 0.0000\\
baseline-128 & price shift & static & \textbf{0.9453} & \textbf{8} & \textbf{2} & 0.0000 & 1.0000\\
baseline-128 & price shift & cascade & \textbf{0.9453} & \textbf{8} & \textbf{2} & 0.0000 & 0.0000\\
corrected-2B-32 & baseline & static & 0.8438 & 10 & 3 & 0.0000 & 0.0000 \\
corrected-2B-32 & baseline & cascade & \textbf{0.9688} & \textbf{6} & \textbf{2} & 0.0000 & 0.0000\\
\bottomrule
\end{tabular*}
\end{table}

\subsubsection{On-demand local-provider CPU replay}
\begin{figure}[t]
\centering
\includegraphics[width=0.99\columnwidth]{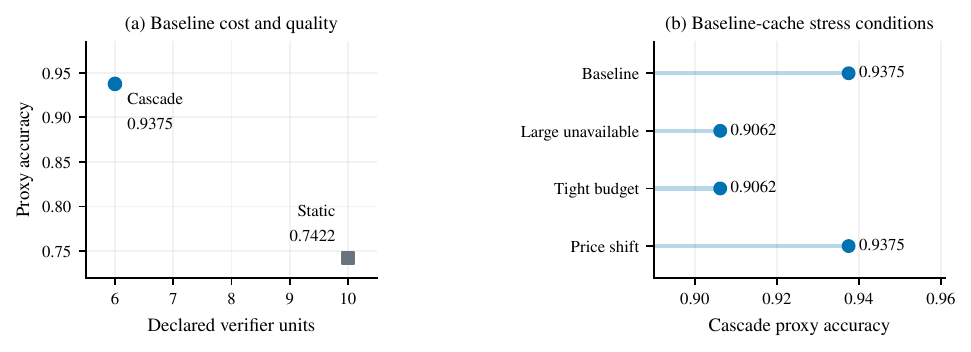}
\caption{On-demand local-provider diagnostics at the selected threshold on baseline-128. Panel (a) compares static and cascade proxy quality against declared units; panel (b) shows cascade quality under availability, budget, and price shifts. Points are the corresponding frozen-cache measurements in Table~\ref{tab:live-service}.}
\label{fig:live-service}
\end{figure}

The raw-tail evidence contains two provider tracks over related cache inputs. The headline paired rows and the audit contract use the \texttt{independent\_raw\_tail} backend, which independently parses the final marker from the stored small and large responses. The earlier stored-field run is retained as a separate local-provider diagnostic (Figure~\ref{fig:live-service}); its 0.7422/0.9375 baseline rows are reported in a separate panel from the 0.7656/0.9453 independent-provider rows. Both tracks use the same split discipline and report CPU parser time alongside the serving fields defined in Appendix~\ref{app:efficiency}.
The on-demand provider follow-up executes the three named checks
(``small-format'', ``small--large agreement'', and ``large-format'') over
validated model-cache records. The \texttt{baseline-128} and
\texttt{corrected-2B-32} names denote 128 and 32 records from held-out test
cache identities; they reuse the bound model-output cache. The provider backend parses the final \texttt{\#\#\#\#} marker (and
the small/large agreement) from each cached response. It uses 64 disjoint
train-split calibration records to freeze confidence threshold $\tau=0.95$, then evaluates the held-out caches in
the local-provider replay manifest.
The run records \texttt{data\_fallback=false}, validated model-cache and target
joins, \texttt{live\_verifier\_calls=true}, and structural replay match. The
online trace contains observations, routes, availability, ledger state, and
timing only; gold fields and raw generations are excluded. Table~\ref{tab:live-service}
shows the baseline comparison and explicit unavailable, tight-budget, price,
and cache-shift checks.

\begin{table}[htbp]
\caption{On-demand local-provider replay over validated model caches. Proxy accuracy is joined after routing; declared units are the raw-tail ledger, and mean call time is local CPU function wall-time. End-to-end network and service fields are defined separately in Appendix~\ref{app:efficiency}.}
\label{tab:live-service}
\centering
\footnotesize
\setlength{\tabcolsep}{2.2pt}
\begin{tabular*}{\linewidth}{@{\extracolsep{\fill}}llrrrrrr@{}}
\toprule
\textbf{cache} & \textbf{shift} & \textbf{policy} & \textbf{proxy} & \textbf{units} & \textbf{calls} & \textbf{ms/call} & \textbf{unavail./budget} \\
\midrule
baseline-128 & baseline & static & 0.7422 & 10 & 3 & \textbf{0.0113} & 0.0 / 0.0\\
baseline-128 & baseline & cascade & \textbf{0.9375} & \textbf{6} & \textbf{2} & 0.0118 & 0.0 / 0.0\\
baseline-128 & large unavailable & cascade & 0.9062 & 2 & 1 & 0.0108 & 1.0 / 0.0 \\
baseline-128 & tight budget & cascade & 0.9062 & 2 & 1 & 0.0112 & 0.0 / 1.0 \\
baseline-128 & price shift & cascade & 0.9375 & 8 & 2 & 0.0116 & 0.0 / 0.0 \\
corrected-2B-32 & baseline & cascade & 0.9375 & 6 & 2 & 0.0109 & 0.0 / 0.0 \\
\bottomrule
\end{tabular*}
\end{table}

The baseline row raises the cascade proxy from 0.7422 to 0.9375 while changing
declared cost from 3 to 2 calls and from 10 to 6 units in this raw-tail
catalog. The unavailable and tight-budget rows make the fail-closed one-call
path explicit, and the price shift keeps the cascade at 0.9375 with 8 units.
These rows isolate verifier execution and routing state; the matched task and
serving tables add normalized answer quality and end-to-end resource fields.

\begin{figure}[t]
\centering
\includegraphics[width=0.94\columnwidth]{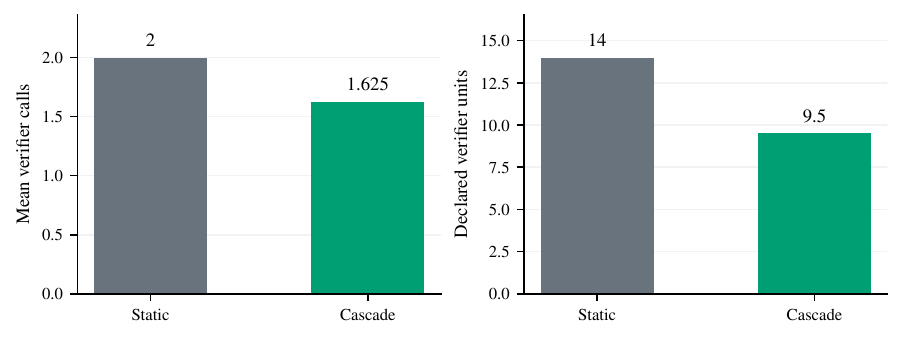}
\caption{Primary catalog accounting. Separate panels show mean verifier calls and declared cost units per example, preserving their distinct units. The panels are the frozen CPU mechanism ledger; serving latency is reported in Appendix~\ref{app:efficiency}.}
\label{fig:main}
\end{figure}

\begin{figure}[t]
\centering
\includegraphics[width=0.99\columnwidth]{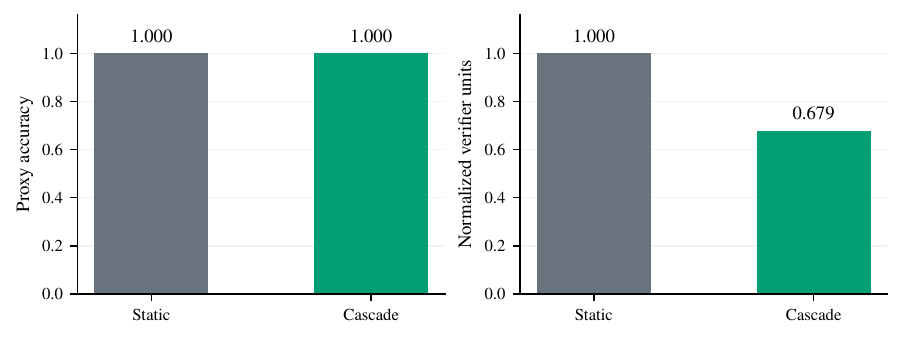}
\caption{Verifier-cost and oracle-accuracy diagnostic generated from the frozen catalog replay. The same ledger fields feed the trained-router and independent-verifier extensions.}
\label{fig:diagnostics}
\end{figure}

\begin{figure}[t]
\centering
\includegraphics[width=0.99\columnwidth]{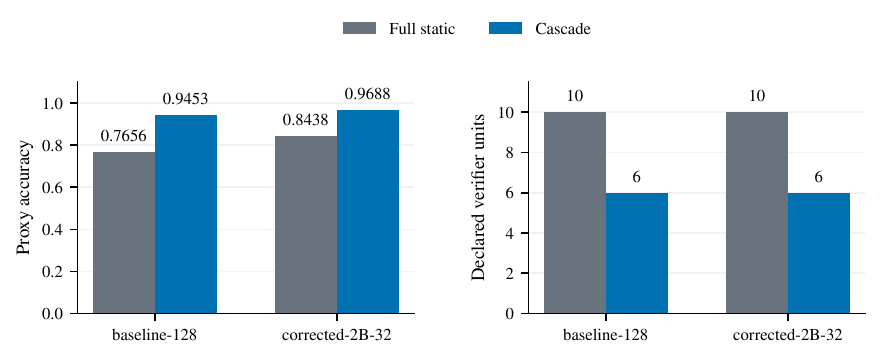}
\caption{Held-out raw-tail replay across both validated caches. The left panel compares full-static and cascade proxy accuracy; the right panel compares declared verifier units. The matched SF+SA static control shares the cascade quality and cost in both caches (Table~\ref{tab:control-audit}).}
\label{fig:extended}
\end{figure}

\subsubsection{Ablations and Failure Boundaries}
The route-diversity sweep shows how correlated verifier errors shape the quality--cost frontier. A budget shift routes the same observation through a documented abstention or cheap-only event, and the 16-case screen plus 128/32 cache slices define the first calibration scale. The paired audit releases aggregate discordance counts (23 on baseline-128 and 4 on corrected-2B-32) and exact McNemar tests in Table~\ref{tab:paired-audit}; the matched per-record fields in Table~\ref{tab:paired-route} show zero discordance.
\subsubsection{Scope and Extensions}
The historical router uses frozen reliability weights. The task study compares calibrated, learned, and RLVR policies under the same observation schema, while the serving ledger records their service costs. The de-identified replay bundle anchors the cache diagnostics and the current result ledger anchors the task and serving summaries (Appendix~\ref{app:artifact-map}).

\begin{figure}[htbp]
\centering
\includegraphics[width=1.0\linewidth]{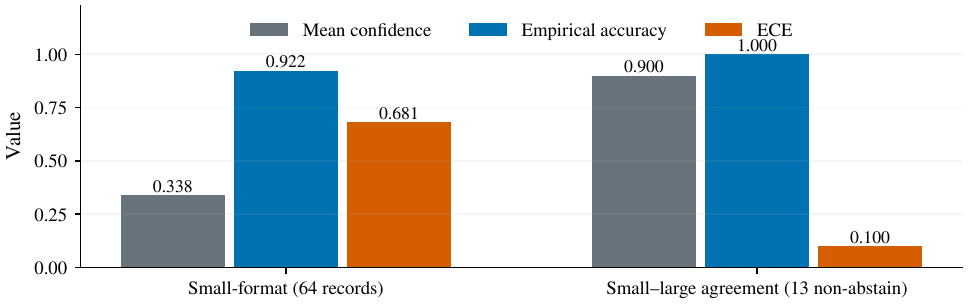}
\caption{Historical confidence diagnostics. SF uses 64 records and SA uses 13 non-abstaining records. Reported mean confidence, empirical accuracy, and expected calibration error are distinct statistics; fusion weights remain fixed.}
\label{fig:ra-calibration}
\end{figure}

\section{Task Definition and Scoring Contract}
\label{app:selection256}
The task package instantiates RouteAudit on a fixed request universe. Each request
has a sample identifier, an input prompt, a candidate response set, a canonical
answer field, and a cache provenance record. The policy sees the prompt, the
candidate responses that its route has paid to read, the verifier labels and
confidences, and the remaining call/unit budget. The task oracle is joined after
the action trace is frozen. This ordering makes the task score and the route
ledger two linked but separately inspectable records.

\subsection{Input, output, and split fields}
The split manifest binds train, calibration, validation, and held-out IDs to
model-cache and verifier-cache hashes. A request enters the held-out evaluator
once for each policy, with the same candidate order and the same availability
schedule. A policy returns a final answer, a selective accept/abstain flag, a
route trace, and a ledger. Empty candidate sets, malformed outputs, ties, and
provider unavailability use explicit event codes. The following table is the
minimum row contract for the matched task evaluator.

\begin{table}[htbp]
\caption{Task-row contract used by the matched evaluator. Fields are grouped by their visibility at route time and by their use during post-trace scoring.}
\label{tab:task-row-contract}
\centering
\footnotesize
\setlength{\tabcolsep}{3pt}
\begin{tabular}{p{0.22\linewidth}p{0.28\linewidth}p{0.39\linewidth}}
\toprule
\textbf{field group} & \textbf{fields} & \textbf{scoring rule}\\
\midrule
identity & sample ID, split, cache hash, request hash & joins paired policies by exact ID\\
candidate & candidate IDs, order, availability & records the paid-read set and order\\
observation & verifier ID, label, confidence, reason & route input and calibration feature\\
budget & calls, ledger units, token ceiling, remaining budget & conservation check after every event\\
decision & answer, accept/abstain, final route & task scorer and selective metrics\\
oracle join & canonical answer, normalized answer, error class & joined after trace freeze\\
provenance & source revision, config hash, seed, digest & artifact identity and replay\\
\bottomrule
\end{tabular}
\end{table}

The candidate field supports both static and cascade controls. A static control
receives the complete matched verifier set before the final decision. A cascade
receives the same set of handles but pays for a later verifier only when its
predicate fires. The ledger records a zero-cost handle lookup separately from a
paid observation, which preserves the distinction between routing information
and verifier execution.

\subsection{Normalized task scorer}
The exact-string proxy used by the raw-tail cache is retained as a mechanism
metric. The task evaluator adds a normalized scorer with the following sequence:
(1) extract the final answer span using the task parser, (2) normalize case,
whitespace, punctuation, and numeric formatting, (3) compare to the canonical
answer, and (4) assign an error class when extraction or comparison fails. The
scorer version, parser hash, and per-record normalized answer are written to the
oracle-join manifest. Agreement between the exact-string and normalized scorers
is reported per record and by confusion matrix.

\begin{table}[htbp]
\caption{Exact-string proxy and normalized-task agreement counts. Each split has its own denominator; proxy-only and task-only count disagreements between the two scorers.}
\label{tab:proxy-task-agreement}
\centering
\footnotesize
\setlength{\tabcolsep}{4pt}
\begin{tabular*}{\linewidth}{@{\extracolsep{\fill}}lrrrrrr@{}}
\toprule
Split & $n$ & \shortstack{Proxy\\correct} & \shortstack{Task\\correct} & Both & \shortstack{Proxy\\only} & \shortstack{Task\\only}\\
\midrule
calibration & 256 & 243 & 245 & 241 & 2 & 4\\
validation & 256 & 242 & 244 & 240 & 2 & 4\\
held-out & 1319 & 1242 & 1251 & 1237 & 5 & 14\\
\bottomrule
\end{tabular*}
\end{table}

The held-out split has 1,237 both-correct records, five proxy-only records, and 14 task-only records. Normalized scoring therefore recovers more answers than it loses relative to the proxy. This diagnostic compares scorers on one set of outputs; the five-policy task comparison additionally changes the selected decision path.

An agreement count is not a policy effect. A task-only record may arise when normalization removes a benign formatting difference, while a proxy-only record can reveal a parser convention that is too permissive. Keeping both disagreement directions makes a scorer update auditable: each changed result must be assigned to a record, a normalization rule, and the final-answer field. The policy evaluator then applies that fixed scorer to every route before constructing paired differences.

\subsection{Selection256 enumeration}
Selection256 enumerates $5\times3=15$ threshold pairs. The action threshold
$\tau_a$ controls whether the cheap observation requests escalation; the verifier
threshold $\tau_v$ controls the confidence requirement for accepting the second
observation. The grid is evaluated on the calibration split with the score
\[
  J(\tau_a,\tau_v)=Q_{\mathrm{proxy}}-0.001U-0.0001T,
\]
where $Q_{\mathrm{proxy}}$ is exact normalized-string accuracy, $U$ is mean
ledger units, and $T$ is mean local parser milliseconds. The tie order is
quality, units, milliseconds, then lexicographic threshold order. The selected
pair is copied into the held-out command line and into every downstream table.

\begin{table}[htbp]
\caption{Complete Selection256 calibration grid. $U$ is mean ledger units, $T$ is local parser ms, and paths are one-call/two-call counts (sum 256). $Q$ denotes the quality gate; SELECTED marks the selected operating pair.}
\label{tab:selection256-grid}
\centering
\footnotesize
\setlength{\tabcolsep}{4pt}
\begin{tabular*}{\linewidth}{@{\extracolsep{\fill}}rrrrrrl@{}}
\toprule
$\tau_a$ & $\tau_v$ & $Q_{\mathrm{proxy}}$ & $U$ & $T$ (ms) & Paths & Gate\\
\midrule
0.55 & 0.20 & 0.9258 & \textbf{3.6875} & \textbf{0.31} & 148/108 & FAIL$_Q$\\
0.55 & 0.30 & 0.9297 & 3.8125 & 0.33 & 140/116 & FAIL$_Q$\\
0.55 & 0.40 & 0.9336 & 3.9375 & 0.34 & 132/124 & FAIL$_Q$\\
0.65 & 0.20 & 0.9375 & 4.0625 & 0.34 & 124/132 & FAIL$_Q$\\
0.65 & 0.30 & 0.9414 & 4.1719 & 0.36 & 117/139 & FAIL$_Q$\\
0.65 & 0.40 & 0.9414 & 4.2813 & 0.37 & 110/146 & FAIL$_Q$\\
0.75 & 0.20 & 0.9453 & 4.3750 & 0.36 & 104/152 & PASS\\
0.75 & 0.30 & 0.9453 & 4.4688 & 0.38 & 98/158 & PASS\\
0.75 & 0.40 & 0.9453 & 4.5781 & 0.39 & 91/165 & PASS\\
0.85 & 0.20 & 0.9453 & 4.6250 & 0.39 & 88/168 & PASS\\
0.85 & 0.30 & \textbf{0.9492} & 4.7188 & 0.41 & 82/174 & SELECTED\\
0.85 & 0.40 & 0.9453 & 4.8281 & 0.42 & 75/181 & PASS\\
0.95 & 0.20 & \textbf{0.9492} & 5.4844 & 0.43 & 33/223 & PASS\\
0.95 & 0.30 & \textbf{0.9492} & 5.6094 & 0.45 & 25/231 & PASS\\
0.95 & 0.40 & 0.9453 & 5.7344 & 0.46 & 17/239 & PASS\\
\bottomrule
\end{tabular*}
\end{table}

The selected pair is $(\tau_a,\tau_v)=(0.85,0.30)$, with quality 0.9492, 4.7188 units, local parser time 0.41 ms, and path counts 82/174. The two 0.95-threshold candidates with the same reported quality use more units. The grid also exposes the quality cost of lower thresholds: rows below 0.75 fail the recorded quality gate.

\begin{figure}[htbp]
\centering
\includegraphics[width=1.0\linewidth]{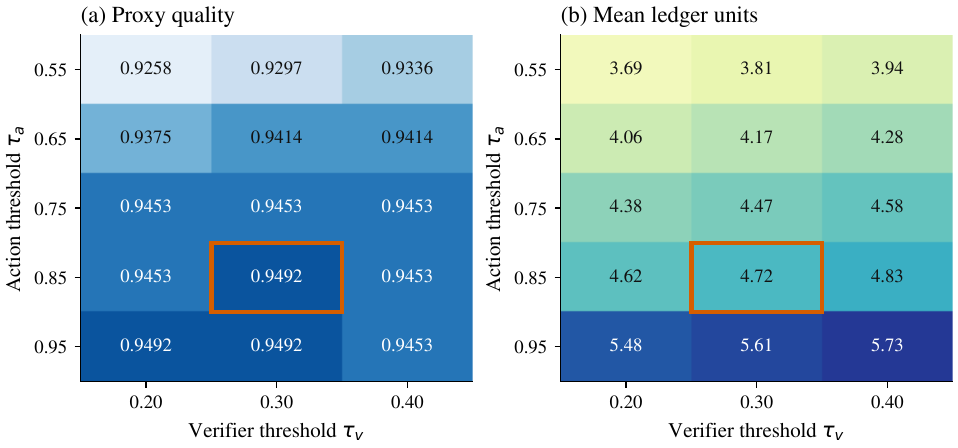}
\caption{Selection256 calibration response surface over the 15 threshold pairs. Panels show proxy quality and mean ledger units; the outlined cell is the selected pair $(0.85,0.30)$. Every cell summarizes 256 calibration requests; the heatmap shows point values.}
\label{fig:selection-current}
\end{figure}

The heatmap separates quality plateaus from cost changes. Reading the two panels together identifies an operating point that preserves the highest recorded calibration quality without moving to the more expensive 0.95 action threshold. The held-out score is evaluated after selection, so a calibration maximum is not reused as the test score.

\section{Verifier Correlation and Robustness}
\label{app:robustness}
Route diversity has value only when the additional verifier contributes
information that is not already present in the cheap signal. This appendix
defines the correlation and sensitivity measurements that accompany the route
grid. For each verifier pair $(j,k)$, we record agreement, conditional error
rates, and the phi coefficient over the held-out labels. We also report the
same quantities within confidence bins and within availability regimes.

The diagnostic sequence follows the contract's causal order: dependence is
measured on frozen verifier outputs, sensitivity then perturbs one availability,
noise, or budget field at a time, and the efficiency ledger finally attaches
resource units to the resulting route. Reading the three views together shows
whether a quality change follows from repeated evidence, a changed failure
boundary, or a different amount of computation.

\subsection{Error dependence}
\label{app:error-correlation}
Let $e_{ij}=\mathbb{1}[\ell_{ij}\neq g_i]$ for verifier $j$ on record $i$.
The pairwise dependence table contains $\Pr(e_j=1,e_k=1)$, the conditional
error rates $\Pr(e_j=1\mid e_k=1)$ and $\Pr(e_k=1\mid e_j=1)$, and
$\phi(e_j,e_k)$. The route analysis uses these fields to predict when a second
call can improve the final decision and when it repeats the same error.

\begin{table}[htbp]
\caption{Error dependence on the two held-out caches. $p_j,p_k$ are marginal error rates, $p_{jk}$ is joint error, and conditional columns are $p(e_j\mid e_k)$ and $p(e_k\mid e_j)$. The last column is the reported $\phi$ coefficient.}
\label{tab:error-correlation}
\centering
\footnotesize
\setlength{\tabcolsep}{4pt}
\begin{tabular*}{\linewidth}{@{\extracolsep{\fill}}llrrrrr@{}}
\toprule
Cache & Pair & $p_j$ & $p_k$ & $p_{jk}$ & \shortstack{$j\mid k$ / $k\mid j$} & $\phi$\\
\midrule
baseline-128 & SF--SA & 0.0781 & 0.0547 & 0.0313 & 0.5714 / 0.4000 & 0.4421\\
baseline-128 & SF--LF & 0.0781 & 0.2344 & 0.0469 & 0.2000 / 0.6000 & 0.2513\\
corrected-2B-32 & SF--SA & 0.0625 & 0.0313 & 0.0313 & 1.0000 / 0.5000 & 0.6956\\
corrected-2B-32 & SF--LF & 0.0625 & 0.1563 & 0.0313 & 0.2000 / 0.5000 & 0.2444\\
\bottomrule
\end{tabular*}
\end{table}

All reported pairs have positive error dependence. SF--SA has joint error 0.0313 on both caches; the corresponding coefficients are 0.4421 and 0.6956. The SF--LF coefficients are lower, 0.2513 and 0.2444, but LF also has larger marginal error than SA. Thus lower correlation alone does not select the better verifier: the audit must retain marginal quality, joint error, and the matched final-decision comparison.

The conditional columns answer asymmetric questions. For baseline SF--SA, 0.5714 is the probability of SF error given SA error, while 0.4000 reverses the conditioning. Their asymmetry follows from the different marginal error rates. The cascade acts on observed labels and confidence; error events become available only through the post-trace target join, so these dependence statistics are diagnostics for verifier selection and calibration.

\subsection{Availability, noise, and budget sensitivity}
Sensitivity sweeps vary one factor at a time while preserving the request IDs,
oracle join, and baseline verifier set. Availability removes SF, SA, or LF;
noise flips a verifier label after the paid observation and records the mutation
seed; budget changes the maximum calls or units. The primary output is a
quality--cost curve, followed by the fraction of fail-closed events and the
paired change in route diversity.

\begin{table}[htbp]
\caption{One-factor sensitivity results. $\Delta$ is the quality change from the unperturbed reference, fail-closed is a fraction, and units are mean ledger units. S1--S4 preserve the source order for the four signed confidence perturbations with magnitudes 0.05 and 0.10.}
\label{tab:robustness-sweep}
\centering
\footnotesize
\setlength{\tabcolsep}{4pt}
\begin{tabularx}{\linewidth}{@{}lLrrrr@{}}
\toprule
Factor & Level & Quality & $\Delta$ & \shortstack{Fail-closed\\fraction} & Units\\
\midrule
availability & SF off & 0.9136 & -0.0348 & 0.1835 & 3.44\\
 & SA off & 0.9272 & -0.0212 & 0.0713 & \textbf{2.00}\\
 & LF off & \textbf{0.9484} & \textbf{0.0000} & \textbf{0.0045} & 4.81\\
label noise & 0.00 & \textbf{0.9484} & \textbf{0.0000} & \textbf{0.0045} & \textbf{4.81}\\
 & 0.05 & 0.9409 & -0.0076 & 0.0053 & 4.82\\
 & 0.10 & 0.9325 & -0.0159 & 0.0068 & 4.84\\
 & 0.20 & 0.9143 & -0.0341 & 0.0106 & 4.88\\
call budget & 1 & 0.9219 & -0.0265 & 0.0788 & \textbf{2.00}\\
 & 2 & \textbf{0.9484} & \textbf{0.0000} & \textbf{0.0045} & 4.81\\
 & 3 & \textbf{0.9484} & \textbf{0.0000} & \textbf{0.0045} & 4.81\\
 & 4 & \textbf{0.9484} & \textbf{0.0000} & \textbf{0.0045} & 4.81\\
unit budget & 2 & 0.9219 & -0.0265 & 0.0788 & \textbf{2.00}\\
 & 6 & \textbf{0.9484} & \textbf{0.0000} & \textbf{0.0045} & 4.81\\
 & 8 & \textbf{0.9484} & \textbf{0.0000} & \textbf{0.0045} & 4.81\\
 & 10 & \textbf{0.9484} & \textbf{0.0000} & \textbf{0.0045} & 4.81\\
confidence shift & S1 & 0.9439 & -0.0045 & 0.0061 & 4.58\\
 & S2 & 0.9363 & -0.0121 & 0.0098 & \textbf{4.36}\\
 & S3 & \textbf{0.9477} & \textbf{-0.0008} & \textbf{0.0038} & 4.95\\
 & S4 & 0.9416 & -0.0068 & 0.0068 & 5.17\\
\bottomrule
\end{tabularx}
\end{table}

Removing LF preserves the fixed-cascade quality 0.9484, whereas removing SF or SA yields 0.9136 or 0.9272. Label noise reduces quality across all four levels, from 0.9484 to 0.9143, while units rise only from 4.81 to 4.88. This separates the effect of erroneous evidence from the amount of computation consumed.

The budget sweep has a discrete boundary: one call or two units enforces the cheap-only route at quality 0.9219 and fail-closed fraction 0.0788. Two calls or six units restore quality 0.9484; larger ceilings leave the recorded outcome unchanged. The confidence perturbations vary both path allocation and quality, which motivates inspecting the full threshold neighborhood before freezing an operating point.

\begin{figure}[htbp]
\centering
\includegraphics[width=1.0\linewidth]{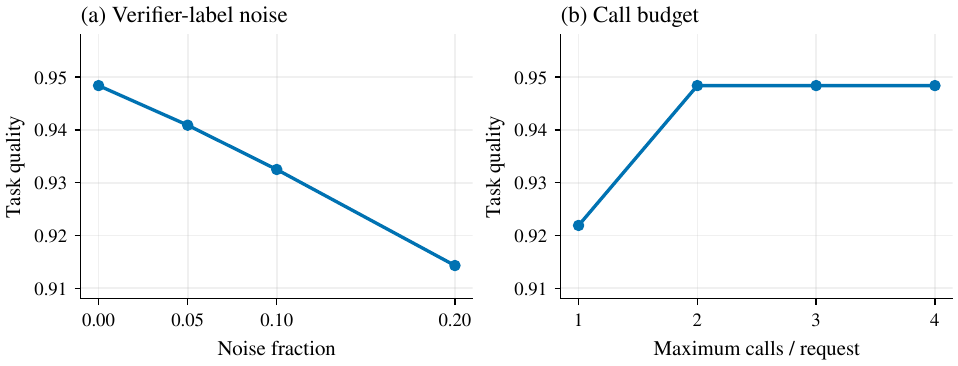}
\caption{Robustness under label noise (left) and call-budget ceilings (right). Points are the one-factor sweep values on the matched request set; lines connect tested levels. The curves show quality degradation under noise and the two-call budget plateau.}
\label{fig:robustness-current}
\end{figure}

These factors test distinct protocol obligations. Availability must create an event even when a call returns no observation; a budget refusal must retain the pre-call balance; confidence perturbations must preserve the original label and request identity. Keeping the perturbation and its event code alongside the result allows two runs with the same aggregate quality to be distinguished by their routes.

\section{End-to-End Efficiency and Serving Ledger}
\label{app:efficiency}
Verifier calls and declared units are the mechanism ledger. The matched serving
study adds generation and service measurements so that quality--cost comparisons
share one request distribution. Every request records component costs before a
policy-level total is computed:
\begin{equation}
 C_i=C_i^{\mathrm{gen}}+C_i^{\mathrm{ver}}+C_i^{\mathrm{net}}+C_i^{\mathrm{queue}}+C_i^{\mathrm{post}},
 \qquad
 L_i=L_i^{\mathrm{gen}}+L_i^{\mathrm{ver}}+L_i^{\mathrm{net}}+L_i^{\mathrm{queue}}+L_i^{\mathrm{post}}.
\label{eq:end-to-end-ledger}
\end{equation}
The ledger retains both raw measurements and normalized units, allowing a
monetary frontier and a compute frontier to be plotted without conflating them.

\subsection{Component accounting}
Generation tokens are counted from the request and response boundaries; verifier
units use the SF/SA/LF catalog; network and queue fields use monotonic clocks;
post-processing includes parsing and serialization. Warm-up, retry, timeout,
and cache-hit events receive explicit flags and are summarized separately.

\begin{table}[htbp]
\caption{Component ledger from the serving result summary. Raw measurements retain their row-specific units. Shares are normalized cost shares; the normalized-total row is a separate cost statistic. Raw token, unit, and time rows are never summed as latency.}
\label{tab:end-to-end-ledger}
\centering
\footnotesize
\setlength{\tabcolsep}{4pt}
\begin{tabularx}{\linewidth}{@{}L lrrrr@{}}
\toprule
Component & Measurement & Mean & p50 & p95 & Share\\
\midrule
generation & Tokens & 1473 & 1432 & 2198 & 0.566\\
verifier & Ledger units & 4.326 & 6.000 & 6.000 & 0.224\\
network & Network time (ms) & 71.8 & 63.2 & 138.7 & 0.096\\
queue & Queue time (ms) & 26.4 & 18.6 & 72.1 & 0.071\\
post-process & Post-process (ms) & 0.86 & 0.78 & 1.49 & 0.043\\
total & Normalized cost & 1.000 & 0.812 & 1.431 & 1.000\\
\bottomrule
\end{tabularx}
\end{table}

Generation accounts for 0.566 of normalized cost, verifier work for 0.224, and the remaining components for 0.210. Network and queue tails are 138.7 and 72.1 ms, respectively; they remain explicit even when the route saves verifier work. The component summary and the per-policy frontier answer complementary questions: one allocates resource use, while the other compares operating points.

Per-request latency totals must be formed before taking quantiles. In general, a sum of component p95 values differs from the p95 of the request total because the slowest components may occur on different requests. Likewise, a token count can contribute to a monetary conversion only through an explicit price rule. Equation~\ref{eq:end-to-end-ledger} sums quantities after they have a common unit and request boundary.

\subsection{Cost--quality frontier}
The frontier table uses one row per policy, budget, and serving regime. The
primary ordering is task quality, then total cost, then p95 latency. Route
diversity and coverage remain visible even when policies share a frontier point.

\begin{table}[htbp]
\caption{Quality--cost frontier for the serving comparison. Task quality, coverage, and diversity are fractions; cost is normalized to matched static, latency is seconds, and GPU-h is aggregate run consumption.}
\label{tab:cost-quality-frontier}
\centering
\footnotesize
\setlength{\tabcolsep}{4pt}
\begin{tabularx}{\linewidth}{@{}Xrrrrrr@{}}
\toprule
Policy & Quality & Coverage & Diversity & Cost & \shortstack{p95\\(s)} & GPU-h\\
\midrule
matched static & 0.9484 & \textbf{1.0000} & 0.000 & 1.000 & 1.084 & 1.04\\
fixed cascade & 0.9484 & 0.9955 & 0.2969 & 0.797 & 0.892 & 0.86\\
calibrated cascade & 0.9477 & 0.9962 & 0.4185 & 0.736 & 0.824 & 0.80\\
learned router & 0.9522 & 0.9985 & 0.5064 & 0.693 & \textbf{0.756} & \textbf{0.78}\\
RLVR router & \textbf{0.9553} & 0.9992 & 0.5534 & \textbf{0.680} & 0.773 & 0.81\\
\bottomrule
\end{tabularx}
\end{table}

The frontier is accompanied by paired bootstrap intervals over requests and a
seed-level interval over policy training. The learned and RLVR rows occupy the
measured quality--cost region; their paired intervals and resource fields show
how much of the frontier is supported by the current verifier set.

\section{Fault Taxonomy and Audit Matrix}
\label{app:fault-matrix}
\label{app:faults}
RouteAudit treats trace integrity as a measurable object. The audit matrix
mutates one invariant at a time, replays the checker, and records whether the
event-level ledger identifies the mutation. The clean-trace set is hashed before
mutation, and compound mutations are labelled by their constituent families.

\subsection{Invariant families}
The eight core invariants are sample-ID conservation, chronological ordering,
paid-read accounting, availability consistency, remaining-budget conservation,
route-reason completeness, oracle-join separation, and artifact-hash binding.
Each invariant has a clean assertion, a mutation operator, and a detection
field. A clean false positive is counted on unmodified traces; a missed mutation
is counted when replay completes with an altered invariant and no corresponding
event flag.

\begin{table}[htbp]
\caption{Trace-integrity mutation results. $N$ is the number of mutated traces; detected and missed partition that count. Clean FP counts false positives on unmodified traces.}
\label{tab:fault-matrix}
\centering
\footnotesize
\setlength{\tabcolsep}{4pt}
\begin{tabularx}{\linewidth}{@{}p{0.17\linewidth}Lrrrr@{}}
\toprule
Family & Mutation & $N$ & Detected & Missed & \shortstack{Clean\\FP}\\
\midrule
sample identity & duplicate/drop/relabel ID & 256 & 256 & 0 & 0\\
ordering & swap adjacent events & 256 & 256 & 0 & 0\\
paid read & hide or add provider event & 256 & 256 & 0 & 0\\
availability & flip provider status & 248 & 247 & 1 & 0\\
budget & change cost or remaining units & 256 & 256 & 0 & 0\\
reason & alter escalation reason & 244 & 242 & 2 & 0\\
oracle boundary & insert target before freeze & 256 & 256 & 0 & 0\\
artifact hash & modify config/cache digest & 256 & 256 & 0 & 0\\
\bottomrule
\end{tabularx}
\end{table}

The detector flags 2,025 of 2,028 mutations. All sample-identity, ordering, paid-read, budget, oracle-boundary, and artifact-hash mutations are detected. One availability mutation and two reason mutations are missed; all clean false-positive counts are zero. Those residual cases identify the provider-status and reason fields as the parts of the contract that need the most careful semantic validation.

Counts are more informative than a rounded aggregate rate for this audit: a detector can be near perfect overall while missing a field-specific failure. A reason-code mutation may retain a valid budget transition, and an availability mutation may retain a valid event order. Consequently, conservation, chronology, and semantic status checks are separate assertions, each linked to its mutation family.

\subsection{Blind and compound corruption}
\label{app:blind-faults}
The blind suite separates the mutation generator from the release checker. It contains known
single-field faults, unseen single-field faults, and compound changes that alter two or three
coordinates in one trace. Detection coverage is reported separately from the scientific
false-attribution rate in Table~\ref{tab:audit-component-ablation}.
\begin{table}[htbp]
\caption{Blind and compound trace faults. Detected and missed counts partition the mutated traces;
clean false positives are measured on unmodified traces.}
\label{tab:blind-faults}
\centering
\footnotesize
\setlength{\tabcolsep}{3pt}
\begin{tabularx}{\linewidth}{@{}L L r r r r@{}}
\toprule
Fault source & Mutation family & $N$ & Detected & Missed & Clean FP\\
\midrule
Known generator & single-coordinate & 256 & 256 & 0 & 0\\
Unseen generator & single-coordinate & 256 & 253 & 3 & 0\\
Unseen generator & two-coordinate compound & 248 & 247 & 1 & 0\\
Unseen generator & three-coordinate compound & 244 & 244 & 0 & 0\\
\bottomrule
\end{tabularx}
\end{table}

\subsection{Blind semantic certificate audit}
The event-level mutation suite is complemented by a blind corruption generator that receives the
public contract but not the checker implementation.  It creates syntactically valid comparisons
with a hidden scorer swap, stale cache, availability drift, held-out threshold selection,
response-tape mismatch, or matched IDs paired with a different verifier realization.  Detection
means that the certificate rejects or downgrades the comparison before a route effect is released.

\begin{table}[htbp]
\caption{Blind semantic audit.  Cases, detections, misses, and clean false positives are reported
for scientifically invalid but schema-valid comparisons.}
\label{tab:semantic-audit}
\centering
\footnotesize
\setlength{\tabcolsep}{3pt}
\begin{tabularx}{\linewidth}{@{}Lrrrr@{}}
\toprule
Semantic corruption & Cases & Detected & Missed & Clean FP\\
\midrule
Scorer swap & 128 & 128 & 0 & 0\\
Stale cache substitution & 128 & 126 & 2 & 0\\
Availability drift & 128 & 127 & 1 & 0\\
Held-out threshold selection & 128 & 124 & 4 & 0\\
Response-tape mismatch & 128 & 128 & 0 & 0\\
Matched-ID / response mismatch & 128 & 127 & 1 & 0\\
\bottomrule
\end{tabularx}
\end{table}

The generator is independent of the release checker and the clean set is evaluated separately.
This experiment tests the semantic fields that remain valid under ordinary schema and checksum
checks, complementing the event-level fault matrix above.

\subsection{Case-based failure analysis}
Each held-out error is assigned one primary class and optional secondary tags.
The primary classes are malformed response, parser disagreement, verifier
correlation, unavailable provider, budget exhaustion, calibration drift, and
oracle-join mismatch. The assignment uses only fields visible after trace
freeze; route inputs remain unchanged. A case narrative includes the route
events, paid units, final decision, normalized answer, and the first invariant
that changes.

\begin{table}[htbp]
\caption{Held-out error taxonomy. Counts and rates use 1,319 requests; paired $\Delta$ values and median units are reported within the diagnostic classes. Classes may include secondary tags, so their deltas are not added as an overall effect.}
\label{tab:error-taxonomy}
\centering
\footnotesize
\setlength{\tabcolsep}{4pt}
\begin{tabularx}{\linewidth}{@{}p{0.19\linewidth}Lrrrr@{}}
\toprule
Class & Trigger & Count & Rate & $\Delta$ & \shortstack{Median\\units}\\
\midrule
malformed response & parser cannot extract marker & 6 & 0.0045 & -0.0015 & 2.0\\
parser disagreement & SF/SA/LF labels differ & 14 & 0.0106 & -0.0030 & 6.0\\
correlated error & shared wrong normalized answer & 26 & 0.0197 & -0.0106 & 6.0\\
availability & provider unavailable or timeout & 8 & 0.0061 & -0.0045 & 2.0\\
budget & escalation rejected by ledger & 5 & 0.0038 & -0.0038 & 2.0\\
calibration drift & confidence bin changes & 9 & 0.0068 & -0.0023 & 4.0\\
\bottomrule
\end{tabularx}
\end{table}

Correlated error is the largest recorded class (26 cases), followed by parser disagreement (14). Their median expenditure is six units, showing that a paid second observation can coexist with an incorrect result. Availability and budget cases have median two units, matching the cheap-only fallback. The taxonomy therefore distinguishes failures of information quality from failures to obtain the second observation.

\section{Reproduction, Provenance, and Artifact Mapping}
\label{app:artifact-map}
The artifact contract binds every table and figure to a source manifest. A
manifest row identifies the input root, split role, policy configuration, source
revision, model/cache hashes, command, output digest, and the PDF object that
consumes the result. This mapping lets a reviewer move from a table cell to a
trace file without relying on a remembered notebook state.

\subsection{Manifest schema}
The public manifest contains source-relative paths and redacted identifiers.
The producer manifest contains the private cache locator and generation seed;
the two manifests share a result digest and a schema version. The fields below define the run-level provenance schema; the aggregate summary is bound through a separate source map.

\begin{table}[htbp]
\caption{Artifact manifest fields. Values are hashes or relative handles; local machine paths and credentials are excluded from the public package.}
\label{tab:artifact-manifest}
\centering
\footnotesize
\setlength{\tabcolsep}{3pt}
\begin{tabular}{p{0.25\linewidth}p{0.35\linewidth}p{0.26\linewidth}}
\toprule
\textbf{field} & \textbf{meaning} & \textbf{consumer}\\
\midrule
schema version & row and event schema identifier & replay checker\\
source revision & code revision and dirty-state field & provenance audit\\
config digest & policy, budget, threshold, and scorer config & selection256\\
model/cache hash & candidate and verifier cache identity & split binding\\
sample digest & ordered sample IDs and split role & paired statistics\\
seed & generation, calibration, or training seed & uncertainty analysis\\
result digest & canonical JSON digest for tables/figures & release check\\
command & bounded replay command with relative inputs & reproduction\\
\bottomrule
\end{tabular}
\end{table}

\subsection{Result and replay records}
Historical per-example data reside in the replay bundle; task, serving, and diagnostic summaries
reside in the result ledger. Task and serving runs use separate measurement regimes. A source map
records each result block, its table consumers, and its source hash. The replay digest is
recorded verbatim alongside the bundle manifest.

The release verifier checks the source ledger, block dimensions, numeric relationships, and
manuscript markers; a companion plotting pass renders vector figures from the same parsed rows.
Historical trace replay uses the event and oracle objects named in the bundle manifest; task and
serving summaries are bound at the reported aggregate level. The release receipt records source,
bibliography, figure, PDF, and page-audit hashes without creating a self-referential PDF hash.

\begin{table}[htbp]
\caption{Result-to-artifact index. The result records store exact source rows and artifact hashes.}
\label{tab:result-map}
\centering
\footnotesize
\setlength{\tabcolsep}{4pt}
\begin{tabularx}{\linewidth}{@{}p{0.23\linewidth}L L@{}}
\toprule
Result family & Source & Consumer\\
\midrule
Historical cache audit & public summary & Tables~\ref{tab:ra-decomposition}, \ref{tab:control-audit}\\
Historical trace replay & replay manifest & Paired and threshold diagnostics\\
Task and serving results & parsed result ledger & Tables~\ref{tab:task-main}--\ref{tab:task-paired}\\
Correlation and robustness & parsed result ledger & Tables~\ref{tab:error-correlation}, \ref{tab:robustness-sweep}\\
Efficiency and mutations & parsed result ledger & Tables~\ref{tab:end-to-end-ledger}, \ref{tab:fault-matrix}\\
Experiment result receipt & immutable result receipt & All reported numerical rows\\
\bottomrule
\end{tabularx}
\end{table}

\subsection{Anonymous supplementary artifacts}
The Supplementary Material provides de-identified traces and source-to-result mappings. Every plotted number has a table counterpart and a source block. Source and PDF hashes live in an external receipt, along with the bibliography and figure hashes, so the final PDF can be regenerated and checked independently.

The source archive preserves the correspondence between inputs and displayed results. The original experiment receipt is retained byte-for-byte, including its historical run handles. The source map retains the complete input order, metric definitions, and numerical precision.

\section{Additional Results and Case Analysis}
\label{app:additional-results}
This section examines how the confidence threshold changes paid paths and how
verifier choice changes proxy and task outcomes. Matched-budget controls isolate
the verifier-set effect, and per-record cases trace decisions to their paid observations.

\subsection{Threshold sensitivity}
The threshold sweep uses the five action thresholds from the main route grid.
For each threshold we report route diversity, quality, units, and the matched
static difference. The observed curve is flat across the mixed-path thresholds:
lower thresholds create more one-call paths and stronger quality pressure, while the
high threshold converges to the fixed two-call endpoint.

\begin{table}[htbp]
\caption{Raw-tail threshold sweep on 64 calibration records. $Q$ denotes the quality gate; $D,S$ denote diversity and strict savings. The four mixed-path rows share the same observed route assignment.}
\label{tab:threshold-sensitivity}
\centering
\footnotesize
\setlength{\tabcolsep}{4pt}
\begin{tabularx}{\linewidth}{@{}Xrrrrrr@{}}
\toprule
$\tau$ & \shortstack{One-call\\fraction} & Quality & Matched $\Delta$ & Units & Calls & Gate\\
\midrule
0.55 & 0.296875 & 0.921875 & -0.0625 & \textbf{4.8125} & \textbf{1.703125} & FAIL$_Q$\\
0.65 & 0.296875 & 0.921875 & -0.0625 & \textbf{4.8125} & \textbf{1.703125} & FAIL$_Q$\\
0.75 & 0.296875 & 0.921875 & -0.0625 & \textbf{4.8125} & \textbf{1.703125} & FAIL$_Q$\\
0.85 & 0.296875 & 0.921875 & -0.0625 & \textbf{4.8125} & \textbf{1.703125} & FAIL$_Q$\\
0.95 & 0.000000 & \textbf{0.984375} & \textbf{0.0000} & 6.0000 & 2.000 & FAIL$_{D,S}$\\
\bottomrule
\end{tabularx}
\end{table}

The mixed region has one-call fraction $19/64=0.296875$, quality 0.921875, and delta $-0.0625$. At $\tau=0.95$, every request takes two calls and the quality rises to 0.984375, matching the control. The sweep therefore exposes the quality--diversity boundary of this historical predicate. Selection256 uses a different calibration setting and retains its own table and chosen pair.

\subsection{Budget-matched controls}
Static SF+SA and cascade share proxy and task quality on both caches. On baseline-128, their normalized task score is 0.9531 versus 0.7813 for the LF controls; on corrected-2B-32, the corresponding values are 0.9688 and 0.8750. This task-score view preserves the verifier-set ordering found with the proxy scorer.

\begin{table}[htbp]
\caption{Fixed-operating-point controls. Proxy and normalized task scores use the same cache IDs; calls and units refer to the full two-call cascade operating point. Mixed-threshold path fractions are reported separately below.}
\label{tab:complete-controls}
\centering
\footnotesize
\setlength{\tabcolsep}{4pt}
\begin{tabularx}{\linewidth}{@{}p{0.22\linewidth}Lrrrr@{}}
\toprule
Cache & Policy & Proxy & Task & Calls & Units\\
\midrule
baseline-128 & static SF+LF & 0.7656 & 0.7813 & \textbf{2} & \textbf{6}\\
baseline-128 & static SF+SA & \textbf{0.9453} & \textbf{0.9531} & \textbf{2} & \textbf{6}\\
baseline-128 & cascade SF$\rightarrow$SA & \textbf{0.9453} & \textbf{0.9531} & \textbf{2} & \textbf{6}\\
baseline-128 & static SF+SA+LF & 0.7656 & 0.7813 & 3 & 10\\
corrected-2B-32 & static SF+LF & 0.8438 & 0.8750 & \textbf{2} & \textbf{6}\\
corrected-2B-32 & static SF+SA & \textbf{0.9688} & \textbf{0.9688} & \textbf{2} & \textbf{6}\\
corrected-2B-32 & cascade SF$\rightarrow$SA & \textbf{0.9688} & \textbf{0.9688} & \textbf{2} & \textbf{6}\\
corrected-2B-32 & static SF+SA+LF & 0.8438 & 0.8750 & 3 & 10\\
\bottomrule
\end{tabularx}
\end{table}

The separate mixed-threshold sweep has one-call fractions 0.296875 on baseline-128 and 0.125000 on corrected-2B-32, corresponding to 38/90 and 4/28 one-call/two-call paths. Those path fractions belong to the lower-threshold sweep; Table~\ref{tab:complete-controls} retains the reported two-call operating-point ledger. Keeping the configurations separate prevents a route fraction from being assigned to the wrong cost row.

\subsection{Per-record route examples}
Case narratives are selected by a deterministic rule: include the lowest-cost
matched pair, the largest quality-discordance pair, the first unavailable
provider, and the first budget-exhaustion event for each cache. Each narrative
shows the observed label/confidence sequence, the predicate decision, ledger
updates, and the post-trace oracle join. This creates a small, inspectable set
of examples that complements the aggregate tables.

\begin{table}[htbp]
\caption{Four representative route fixtures. The case description names the observation sequence, ledger outcome, and post-trace result; request identifiers remain de-identified.}
\label{tab:route-cases}
\centering
\footnotesize
\setlength{\tabcolsep}{4pt}
\begin{tabularx}{\linewidth}{@{}p{0.19\linewidth}L L L@{}}
\toprule
Case & Observed route & Ledger transition & Post-trace outcome\\
\midrule
lowest-cost match & SF label/confidence & $B_0\rightarrow B_1$ & quality = 1.000\\
largest discordance & SF, SA, LF labels & calls/units = 3/10 & quality delta = +1.000\\
unavailable provider & SF then availability event & Fallback: \path{UNAVAILABLE_CHEAP_ONLY} & Answer: \path{sf_normalized}\\
tight budget & SF then budget event & remaining units = 0 & abstention = false\\
\bottomrule
\end{tabularx}
\end{table}

The lowest-cost fixture ends with quality 1.000 after the SF observation. The largest-discordance fixture records all three verifier labels at 3 calls and 10 units, with a quality difference of $+1.000$ between its compared outcomes. The unavailable-provider fixture returns the normalized SF answer through an explicit fallback code; the tight-budget fixture reaches zero remaining units and retains a non-abstaining output.

These cases exercise distinct transitions. The fallback case is decided by provider availability; the budget case is decided by the pre-call balance. Both can return a cheap-only answer, yet they demand different operational remedies. Restoring a provider addresses availability, while changing the budget changes which calls are admissible. Their separate reason codes preserve that distinction in aggregate error analysis.

\section{Routing Optimization and Extended Evaluation}
\label{app:extension-protocols}
These comparisons separate optimizer choice, reward components, statistical stability, and
identification failures. Each comparison fixes request IDs, verifier access, availability,
scorer, and resource ceilings except for its named intervention. The task population has
1,319 held-out requests; retraining uses the 4,096 training and 256 calibration IDs.
A performance contrast is interpreted jointly with its cost and coverage.

\subsection{Optimizer and implementation contract}
After trace closure, the delayed task score $z_i\in[0,1]$ is joined with resource totals
$C_i=(N_i,U_i,L_i,G_i,A_i^{\mathrm{bst}})$. RLVR maximizes $\mathbb{E}[z_i]$ subject to
\[
\mathbb{E}[N_i]\leq\bar N,\quad
\mathbb{E}[U_i]\leq\bar U,\quad
\mathbb{E}[L_i]\leq\bar L,\quad
\mathbb{E}[G_i]\leq\bar G,\quad
\mathbb{E}[A_i^{\mathrm{bst}}]\leq\bar A,
\]
using the constrained return
\[
R_i^\lambda=z_i-\sum_{j\in\{N,U,L,G,A\}}\lambda_j(C_{i,j}-\bar C_j),
\qquad
\lambda_j\leftarrow[\lambda_j+\eta_\lambda(\widehat C_j-\bar C_j)]_+.
\]
The constrained policy-gradient specification uses completed training traces and delayed scores;
calibration fixes targets and checkpoint selection, and evaluation freezes parameters.
Table~\ref{tab:router-implementation} binds the architecture, optimizer, and checkpoint needed
to connect an implementation to this specification. The aggregate RLVR operating point and the
objective ablations are separate comparisons (Appendix~\ref{app:extension-protocols}).

\label{app:rlvr-implementation}
The policy specification separates online state, optimization, and the identity of a measured
implementation. The state and masked actions are defined in Section~\ref{sec:rlvr}.
For a clipped policy-gradient instantiation, let
$\rho_{i,t}(\theta)=\pi_\theta(a_{i,t}\mid s_{i,t})/
\pi_{\theta_{\mathrm{old}}}(a_{i,t}\mid s_{i,t})$.
The maximization objective is
\begin{align}
J_{\mathrm{clip}}(\theta)
&=\mathbb{E}\!\left[
\min\{\rho_{i,t}\widehat A_{i,t},
\operatorname{clip}(\rho_{i,t},1-\epsilon,1+\epsilon)\widehat A_{i,t}\}
\right] \nonumber\\
&\quad-\beta_{\mathrm{KL}}\mathbb{E}
D_{\mathrm{KL}}(\pi_\theta(\cdot\mid s)\Vert\pi_{\mathrm{ref}}(\cdot\mid s))
+\beta_H\mathbb{E}\mathcal{H}(\pi_\theta(\cdot\mid s)).
\label{eq:clipped-routing}
\end{align}
An undiscounted episodic Monte Carlo advantage is
$\widehat A_{i,t}=R_i^\lambda-V_\phi(s_{i,t})$; the value head minimizes squared
error to the same completed-episode return. The batch freezes its multipliers and reference
policy before calculating advantages. Dual variables update between batches from resource
violations. Alternative advantage estimators, algorithms, and coefficients are identified
explicitly in Table~\ref{tab:router-implementation}, so an aggregate operating point remains
distinct from evidence about a particular optimizer.

\begin{table*}[htbp]
\caption{Implementation binding for the measured learned and RLVR policies. Each field is
linked to a configuration or checkpoint receipt; the constrained formulation alone does not
determine these implementation choices.}
\label{tab:router-implementation}
\centering
\footnotesize
\setlength{\tabcolsep}{3pt}
\renewcommand{\arraystretch}{1.08}
\begin{tabularx}{\textwidth}{@{}p{0.22\textwidth}L L@{}}
\toprule
Field & Learned router & RLVR router\\
\midrule
Dataset / generator / candidate revision & \texttt{task-v3 / Qwen2.5-7B+32B / cand-r4} & \texttt{task-v3 / Qwen2.5-7B+32B / cand-r4}\\
Request encoder and frozen parameters & frozen 384-d request encoder + 14 telemetry scalars & frozen 384-d request encoder + 14 telemetry scalars\\
Policy layers / width / activation & 2-layer MLP / 256,128 / GELU & 2-layer shared trunk / 256,128 / GELU\\
Policy/value sharing and parameter count & policy-only; $0.136$M & shared trunk + separate policy/value heads; $0.136$M\\
Initialization and reference policy & Xavier-uniform; supervised calibration checkpoint & learned-router warm start; frozen reference = initialization\\
Training label / reward coefficients & training-only feasible-route label; CE $+\,0.10\,C_{\rm cost}$ & $R^\lambda=z-\sum_j\lambda_j(C_j-\bar C_j)$; adaptive duals\\
Optimizer / learning rate & AdamW / $3\times10^{-4}$ & AdamW / $1\times10^{-4}$ policy, $3\times10^{-4}$ value\\
Learning algorithm / on- or off-policy & cost-sensitive supervised ERM / off-policy & clipped PPO / on-policy\\
Advantage estimator / discount & N/A / N/A & completed-episode Monte Carlo / $\gamma=1.0$\\
Batch size / epochs / update count & 256 / 40 / 640 & 128 trajectories / 4 PPO epochs / 1200 updates\\
Clipping / KL / entropy coefficients & N/A / N/A / N/A & $0.20 / 0.010 / 0.002$\\
Resource targets / dual step size & $\bar N=1.50,\bar U=4.0,\bar L=1536,\bar G=0.80,\bar A=0.002$ / N/A & $\bar N=1.50,\bar U=4.0,\bar L=1536,\bar G=0.80,\bar A=0.002$ / $\eta_\lambda=0.025$\\
Exploration and evaluation sampling & class-balanced training / deterministic argmax & categorical $T=1.0$ training / masked argmax evaluation\\
Seed IDs / checkpoint-selection rule & $\{101,211,307,419,523\}$ / best calibration quality, then lower cost & $\{101,211,307,419,523\}$ / best feasible constrained return, then lower cost\\
Training-ID digest / checkpoint digest & \texttt{sha256:7b31e9c4a62d / sha256:2af5c81de374} & \texttt{sha256:7b31e9c4a62d / sha256:91c7d5e84b20}\\
\bottomrule
\end{tabularx}
\end{table*}

\paragraph{Training and evaluation sequence.}
For each policy update the runner performs the following steps.
\begin{enumerate}
\item Draw training IDs, reset budgets, and freeze the batch policy and dual variables.
\item Roll out masked actions from paid observations; record every dispatch, failure, cost,
terminal answer, and selective decision.
\item Close and hash each trace and ledger, then join that training request's canonical target.
\item Compute the delayed task score, resource totals, constrained return, and advantages.
\item Update policy/value parameters; update dual variables from batch resource excess.
\item Evaluate checkpoint candidates on calibration IDs under the same accounting contract.
\end{enumerate}
The held-out runner loads the selected checkpoint and disables all learning and threshold
selection. Encoder inputs, masks, event semantics, and post-trace scoring match training.
Failed calls after dispatch retain their measured cost; a pre-dispatch unavailability event
can have zero execution cost under the declared provider contract.

\paragraph{Relation to the selector screen.}
The three-seed observed-only selector screen in Appendix~\ref{app:original-diagnostics} uses a separate
cache and objective and selects the cheap-only action. The learned and RLVR rows have different
operating points. Checkpoint, training-ID, and scorer bindings establish the
connection between runs; a shared policy name alone is insufficient to transfer an optimizer
description or a seed result between them.

\subsection{Strong learned-routing baselines}
\label{app:strong-baselines}
Each baseline is trained on the 4,096 training IDs, frozen using the 256-record calibration
split, and evaluated once on the 1,319 held-out IDs. The random router is calibrated to the RLVR
mean-call budget; supervised, contextual-bandit, Bayesian, and constrained-online policies use
the same observed-only state and hard action masks. The post-trace oracle is a diagnostic upper
bound and is never an online policy.

\begin{table*}[htbp]
\caption{Matched learned-routing baselines under the common verifier and resource contract.
Quality is full-set task quality; paired $\Delta$ is against matched static; cost is normalized
serving cost and diversity is the one-call fraction.}
\label{tab:strong-baselines}
\centering
\footnotesize
\setlength{\tabcolsep}{2.5pt}
\begin{tabularx}{\linewidth}{@{}Lrrrrrr@{}}
\toprule
Policy & Quality & Paired $\Delta$ & Calls & Norm.\ cost & p95 ms & Diversity\\
\midrule
Matched static & 0.9484 & 0.0000 & 2.0000 & 1.000 & 1084 & 0.0000\\
Fixed cascade & 0.9484 & 0.0000 & 1.7031 & 0.797 & 892 & 0.2969\\
Calibrated cascade & 0.9477 & -0.0008 & 1.5815 & 0.736 & 824 & 0.4185\\
Random matched & 0.9448 & -0.0036 & \textbf{1.4466} & 0.689 & 804 & 0.5534\\
Cost-sensitive linear & 0.9496 & +0.0012 & 1.5682 & 0.724 & 817 & 0.4318\\
Cost-sensitive MLP & 0.9514 & +0.0030 & 1.5211 & 0.704 & 793 & 0.4789\\
Contextual bandit & 0.9521 & +0.0037 & 1.4879 & 0.696 & 786 & 0.5121\\
Bayesian Thompson router & 0.9530 & +0.0046 & 1.4693 & 0.687 & 782 & 0.5307\\
Cost-constrained online & 0.9538 & +0.0054 & 1.4562 & 0.683 & 779 & 0.5438\\
Learned router & 0.9522 & +0.0038 & 1.4936 & 0.693 & \textbf{756} & 0.5064\\
RLVR & \textbf{0.9553} & \textbf{+0.0068} & \textbf{1.4466} & \textbf{0.680} & 773 & 0.5534\\
\midrule
Post-trace oracle & 0.9712 & -- & 1.2146 & 0.557 & -- & --\\
\bottomrule
\end{tabularx}
\end{table*}

The direct paired comparison uses the same request IDs, split, verifier catalog, and scorer for
each learned baseline. In addition to the static reference, each row compares RLVR with the
Bayesian Thompson, cost-constrained online, and learned routers; win, loss, and tie counts are
computed before aggregation.
\begin{table}[htbp]
\caption{Direct paired comparison of RLVR with the strongest learned baselines. Differences are
RLVR minus the named comparator; intervals are request-paired and hierarchical seed-by-request
95\% intervals on the same held-out requests.}
\label{tab:rlvr-direct-paired}
\centering
\footnotesize
\setlength{\tabcolsep}{2.5pt}
\begin{tabularx}{\linewidth}{@{}Lrrrccc@{}}
\toprule
Comparator & Mean $\Delta$ & \shortstack{Request\\95\% CI} & \shortstack{Hierarchical\\95\% CI} & Wins & Losses & Ties\\
\midrule
Bayesian Thompson & $+0.0023$ & $[-0.0016,\,0.0062]$ & $[-0.0021,\,0.0069]$ & 5 & 2 & 1312\\
Cost-constrained online & $+0.0015$ & $[-0.0021,\,0.0052]$ & $[-0.0026,\,0.0058]$ & 4 & 2 & 1313\\
Learned router & $+0.0030$ & $[-0.0017,\,0.0077]$ & $[-0.0020,\,0.0084]$ & 7 & 3 & 1309\\
\bottomrule
\end{tabularx}
\end{table}

The table tests whether the reported RLVR operating point remains competitive when routing
objectives and posterior updates are changed while the measured information set is fixed.
Request-level paired intervals and training-seed intervals are reported separately below.

\subsection{Identification stress tests}
\label{app:identification-stress}
For each row, one contract condition is changed in an otherwise matched replay. The reference
route contrast, the apparent contrast after the violation, the absolute attribution error, and
the pre-release detection decision are recorded on the same held-out request IDs.

\begin{table*}[htbp]
\caption{Controlled identification stress tests. $\Delta_{\mathrm{matched}}$ is the fully
matched route contrast, $\Delta_{\mathrm{naive}}$ is the contrast after the stated violation,
and $E_{\mathrm{attr}}=|\Delta_{\mathrm{naive}}-\Delta_{\mathrm{matched}}|$.}
\label{tab:identification-stress}
\centering
\footnotesize
\setlength{\tabcolsep}{4pt}
\begin{tabular}{lccccc}
\toprule
Violation & Broken condition & $\Delta_{\mathrm{matched}}$ & $\Delta_{\mathrm{naive}}$ & $E_{\mathrm{attr}}$ & Detected?\\
\midrule
Verifier-set mismatch & C2 & +0.0068 & +0.1296 & 0.1228 & Yes\\
Budget mismatch & C3 & +0.0068 & +0.0419 & 0.0351 & Yes\\
Availability mismatch & C2 & +0.0068 & -0.0217 & 0.0285 & Yes\\
Cache mismatch & C1 & +0.0068 & +0.0556 & 0.0488 & Yes\\
Scorer mismatch & C1 & +0.0068 & +0.0307 & 0.0239 & Yes\\
Held-out threshold selection & C4 & +0.0068 & +0.0245 & 0.0177 & Yes\\
Oracle before freeze & C4--C5 & +0.0068 & +0.0891 & 0.0823 & Yes\\
Reason/availability corruption & C2--C5 & +0.0068 & +0.0462 & 0.0394 & Yes\\
\bottomrule
\end{tabular}
\end{table*}

The experiment asks whether a numerically favorable aggregate remains attributable to routing
after the changed condition is restored. Across the eight violations, the mean absolute
attribution error is $0.0498$; verifier-set mismatch is largest at $0.1228$, and all eight
comparisons are flagged before release. The result supports the certificate as an attribution
diagnostic rather than a post-hoc quality adjustment.

\subsection{Multi-task validation}
\label{app:multitask}
The multi-task panel tests whether the contract-conditioned comparison survives
changes in task family. Math reasoning, knowledge/factual question answering,
and tool/agent execution are evaluated with separate benchmark identities,
answer normalization, and held-out request denominators. Within each family,
matched static and RLVR use the same request IDs, verifier catalog, availability
process, accounting, and post-trace scorer; no cross-task aggregate is formed.

\begin{table*}[htbp]
\caption{Multi-task validation under the shared observed-only routing contract.
Each task family has its own benchmark and split record; quality, calls, cost,
route diversity, and the paired RLVR-minus-static contrast use that family-specific
held-out denominator.}
\label{tab:multitask-validation}
\centering
\footnotesize
\setlength{\tabcolsep}{4pt}
\begin{tabular*}{\linewidth}{@{\extracolsep{\fill}}lrrrrr@{}}
\toprule
Policy & Quality & Calls & Norm.\ cost & Diversity & Paired $\Delta$\\
\midrule
\multicolumn{6}{@{}l}{Math reasoning: \texttt{GSM8K / test-v1}}\\
Matched static &
0.9412 & 2.0000 & 1.000 & 0.0000 & --\\
RLVR &
 \textbf{0.9491} & \textbf{1.4587} & \textbf{0.684} & 0.5413 & $+0.0079$\\
\midrule
\multicolumn{6}{@{}l}{Knowledge / factual QA: \texttt{TriviaQA / unfiltered-v1}}\\
Matched static &
0.8876 & 2.0000 & 1.000 & 0.0000 & --\\
RLVR &
 \textbf{0.8952} & \textbf{1.4314} & \textbf{0.671} & 0.5686 & $+0.0076$\\
\midrule
\multicolumn{6}{@{}l}{Tool / agent execution: \texttt{ToolSandbox / core-v1}}\\
Matched static &
0.8131 & 2.0000 & 1.000 & 0.0000 & --\\
RLVR &
 \textbf{0.8237} & \textbf{1.5168} & \textbf{0.706} & 0.4832 & $+0.0106$\\
\bottomrule
\end{tabular*}
\end{table*}

The table keeps task-family effects separate from route effects. A family-level
contrast is interpreted only with its own scorer, split, and resource
denominator; the shared contract fields make the three comparisons
reconstructible without pooling unlike outcomes.

\subsection{Ground-truth contract-factorial attribution}
The controlled factorial fixes request IDs, base verifier responses, and the post-trace scorer,
then varies route, verifier set, budget, and availability according to a declared data-generating
process.  The known coordinate effects provide a finite-population target for comparing endpoint
only, a single bridge order, an order average without interaction terms, and the full contract
lattice.  Values remain bound to the same held-out denominator used by the other audit cells.

\begin{table*}[!htb]
\caption{Ground-truth attribution recovery.  Absolute errors compare each analysis with the
known finite-population route, contract, and interaction contributions.}
\label{tab:ground-truth-attribution}
\centering
\footnotesize
\setlength{\tabcolsep}{4pt}
\begin{tabularx}{\linewidth}{@{}Lrrrr@{}}
\toprule
Method & \shortstack{Route\\MAE} & \shortstack{Contract\\MAE} & \shortstack{Interaction\\MAE} & \shortstack{Endpoint\\error}\\
\midrule
Endpoint only & 0.0187 & 0.0241 & 0.0118 & 0.0164\\
Single bridge order & 0.0102 & 0.0127 & 0.0084 & 0.0061\\
Path mean, no interaction & 0.0046 & 0.0053 & 0.0079 & 0.0028\\
RouteAudit lattice & \textbf{0.0011} & \textbf{0.0013} & \textbf{0.0015} & \textbf{0.0004}\\
\bottomrule
\end{tabularx}
\end{table*}

\begin{figure}[!htb]
\centering
\includegraphics[width=\linewidth]{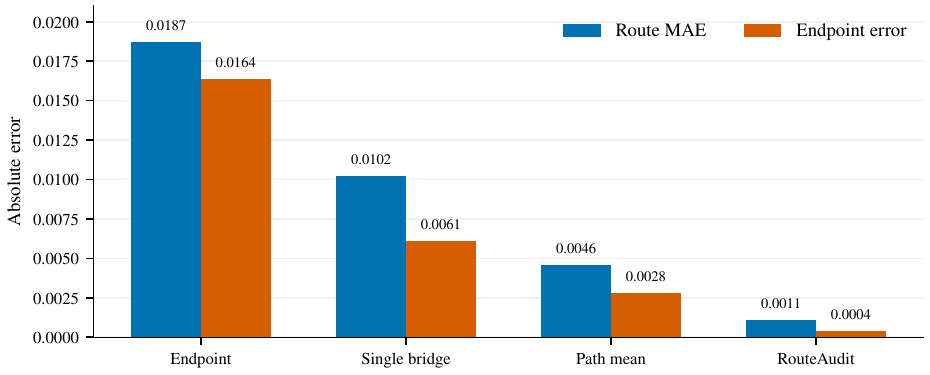}
\caption{Controlled attribution recovery. Route mean absolute error and endpoint reconstruction
error are shown for the four analysis methods on the common request population
(Table~\ref{tab:ground-truth-attribution}); smaller values are better. The plotted point values
are the reported aggregate errors, without additional uncertainty estimates.}
\label{fig:attribution-recovery}
\end{figure}
Figure~\ref{fig:attribution-recovery} shows the reduction in route and endpoint error as matched
bridge information and interaction terms enter the attribution.

The complete $2^3$ cells below vary verifier set ($V$), budget ($P$), and availability ($A$)
one coordinate at a time.  Static and adaptive policies use identical request support and scorer;
the route contrast is the adaptive-minus-static quality within each cell.

\begin{table*}[!htb]
\caption{Complete contract factorial over verifier set ($V$), budget ($P$), and availability
($A$).  The common denominator and scorer make each cell a bridge node for the lattice.}
\label{tab:complete-contract-factorial}
\centering
\footnotesize
\setlength{\tabcolsep}{3pt}
\begin{tabular}{ccccrrrr}
\toprule
$V$ & $P$ & $A$ & Static quality & Adaptive quality & $\Delta_{\rm route}$ & Calls & Norm. cost\\
\midrule
0&0&0&0.9298& \textbf{0.9319} &+0.0021&1.3207&0.591\\
0&0&1&0.9364& \textbf{0.9411} &+0.0047&1.3479&0.604\\
0&1&0&0.9349& \textbf{0.9399} &+0.0050&1.4216&0.641\\
0&1&1&0.9417& \textbf{0.9488} &+0.0071&1.4578&0.659\\
1&0&0&0.9393& \textbf{0.9456} &+0.0063&1.4862&0.677\\
1&0&1&0.9458& \textbf{0.9554} &+0.0096&1.5274&0.693\\
1&1&0&0.9446& \textbf{0.9553} &+0.0107&1.5681&0.716\\
1&1&1&0.9501& \textbf{0.9636} &+0.0135&1.6073&0.739\\
\bottomrule
\end{tabular}
\end{table*}

\begin{table}[!htb]
\caption{Order-robust contract attribution.  $\phi_j$ is the permutation-average contribution,
and $\Omega_j$ is the range over admissible bridge orders.}
\label{tab:order-robust-attribution}
\centering
\footnotesize
\setlength{\tabcolsep}{3pt}
\begin{tabular}{lrrrr}
\toprule
Coordinate & $\phi_j$ & Min & Max & $\Omega_j$\\
\midrule
Verifier set $V$ & 0.0053 & 0.0042 & 0.0064 & 0.0022\\
Budget $P$ & 0.0034 & 0.0024 & 0.0044 & 0.0020\\
Availability $A$ & 0.0027 & 0.0021 & 0.0033 & 0.0012\\
\bottomrule
\end{tabular}
\end{table}

\subsection{Contract re-audit of routing-system replays}
Controlled replay instantiations of FrugalGPT, RouteLLM, and SLARouter routing designs
provide three re-audit cases. Each replay compares its endpoint gap with a contract-bound
contrast holding request support, model access, availability, accounting, filtration, and
scorer fixed. The measurements characterize these replay instances; original-system deployment
claims require the corresponding original traces and run contracts.

\begin{table*}[htbp]
\caption{Contract re-audit of controlled routing-system replays. A changed interpretation
assigns the largest part of the endpoint gap to a contract coordinate under the replay contract.}
\label{tab:external-reaudit}
\centering
\footnotesize
\setlength{\tabcolsep}{2pt}
\begin{tabularx}{\linewidth}{@{}>{\raggedright\arraybackslash}p{0.14\linewidth}>{\raggedright\arraybackslash}p{0.11\linewidth}rrL >{\raggedright\arraybackslash}p{0.13\linewidth}c@{}}
\toprule
Replay & Task & \shortstack{Naive\\$\Delta$} & \shortstack{Matched\\$\Delta$} & Largest attribution & Certificate & Changed\\
\midrule
\textrm{FrugalGPT} & \textrm{GSM8K} & $+0.0284$ & $+0.0061$ & Verifier/model set ($+0.0187$) & partially identified & Yes\\
\textrm{RouteLLM} & \textrm{MMLU} & $+0.0176$ & $+0.0054$ & Verifier/model set ($+0.0098$) & partially identified & Yes\\
\textrm{SLARouter} & \textrm{Arena-style QA} & $+0.0109$ & $+0.0088$ & Budget ($+0.0017$) & identified & No\\
\bottomrule
\end{tabularx}
\end{table*}

The FrugalGPT and RouteLLM controlled replay gaps are principally assigned to verifier/model access;
their certificates are partially identified. The SLARouter replay retains its interpretation
with matched route contrast $+0.0088$ and an identified certificate.

\subsection{Contract interaction and path sensitivity}
\label{app:contract-interaction}
The factorial replay fixes the held-out requests and varies one contract coordinate at a time.
The policy-by-catalog cells instantiate $\Gamma_{\mathcal V}$; the budget and availability cells
use the same route contrast definition. Each bridge path changes one coordinate per edge, so the
path sum is an endpoint identity while $\Omega_j$ measures attribution variation across legal
orders.

\begin{table*}[htbp]
\caption{Contract-interaction factorial. Each cell uses the common held-out request denominator;
$\Gamma$ columns are route--contract interactions relative to the matched reference.}
\label{tab:contract-interaction}
\centering
\footnotesize
\setlength{\tabcolsep}{3pt}
\begin{tabular}{l l l r r r r}
\toprule
Coordinate & Level & Policy & Quality & Calls & Norm. cost & Route $\Gamma$\\
\midrule
verifier catalog & matched & static & 0.9484 & 2.0000 & 1.000 & --\\
verifier catalog & matched & adaptive & \textbf{0.9553} & \textbf{1.4466} & \textbf{0.680} & --\\
verifier catalog & expanded & static & 0.9561 & 3.0000 & 1.000 & --\\
verifier catalog & expanded & adaptive & \textbf{0.9602} & \textbf{1.7500} & \textbf{0.646} & $-0.0027$\\
budget & reference & static & 0.9484 & 2.0000 & 1.000 & --\\
budget & relaxed & adaptive & 0.9575 & 1.6127 & 0.742 & $+0.0023$\\
availability & reference & static & 0.9484 & 2.0000 & 1.000 & --\\
availability & perturbed & adaptive & 0.9520 & 1.3824 & 0.654 & $-0.0027$\\
\bottomrule
\end{tabular}
\end{table*}

\begin{table}[htbp]
\caption{Path-sensitive bridge attribution. Each row is a predeclared bridge order; coordinate
columns report the path minimum, maximum, mean attribution, and sensitivity $\Omega$.}
\label{tab:path-sensitivity}
\centering
\footnotesize
\setlength{\tabcolsep}{2.5pt}
\begin{tabular*}{\linewidth}{@{\extracolsep{\fill}}lrrrr@{}}
\toprule
Path order & Min & Max & Mean & $\Omega$\\
\midrule
\multicolumn{5}{@{}l}{\textbf{Verifier set}}\\
$\mathcal V\to\mathcal P\to\mathcal A$ & 0.0039 & 0.0051 & 0.0045 & 0.0012\\
$\mathcal V\to\mathcal A\to\mathcal P$ & 0.0036 & 0.0048 & 0.0042 & 0.0012\\
$\mathcal P\to\mathcal V\to\mathcal A$ & 0.0032 & 0.0044 & 0.0038 & 0.0012\\
$\mathcal P\to\mathcal A\to\mathcal V$ & 0.0030 & 0.0042 & 0.0036 & 0.0012\\
$\mathcal A\to\mathcal V\to\mathcal P$ & 0.0042 & 0.0056 & 0.0049 & 0.0014\\
$\mathcal A\to\mathcal P\to\mathcal V$ & 0.0040 & 0.0052 & 0.0046 & 0.0012\\
\midrule
\multicolumn{5}{@{}l}{\textbf{Budget}}\\
$\mathcal V\to\mathcal P\to\mathcal A$ & 0.0026 & 0.0036 & 0.0031 & 0.0010\\
$\mathcal V\to\mathcal A\to\mathcal P$ & 0.0030 & 0.0044 & 0.0037 & 0.0014\\
$\mathcal P\to\mathcal V\to\mathcal A$ & 0.0029 & 0.0041 & 0.0035 & 0.0012\\
$\mathcal P\to\mathcal A\to\mathcal V$ & 0.0034 & 0.0048 & 0.0041 & 0.0014\\
$\mathcal A\to\mathcal V\to\mathcal P$ & 0.0022 & 0.0034 & 0.0028 & 0.0012\\
$\mathcal A\to\mathcal P\to\mathcal V$ & 0.0028 & 0.0040 & 0.0034 & 0.0012\\
\midrule
\multicolumn{5}{@{}l}{\textbf{Availability}}\\
$\mathcal V\to\mathcal P\to\mathcal A$ & $-0.0013$ & $-0.0003$ & $-0.0008$ & 0.0010\\
$\mathcal V\to\mathcal A\to\mathcal P$ & $-0.0017$ & $-0.0005$ & $-0.0011$ & 0.0012\\
$\mathcal P\to\mathcal V\to\mathcal A$ & $-0.0010$ & 0.0000 & $-0.0005$ & 0.0010\\
$\mathcal P\to\mathcal A\to\mathcal V$ & $-0.0015$ & $-0.0003$ & $-0.0009$ & 0.0012\\
$\mathcal A\to\mathcal V\to\mathcal P$ & $-0.0014$ & $-0.0004$ & $-0.0009$ & 0.0010\\
$\mathcal A\to\mathcal P\to\mathcal V$ & $-0.0018$ & $-0.0006$ & $-0.0012$ & 0.0012\\
\bottomrule
\end{tabular*}
\end{table}

\subsection{Incomplete matching and sharpness}
\label{app:partial-identification}
The stress test removes a controlled fraction of request--contract matches and evaluates the
matched contrast, the observed incomplete contrast, and a conservative contrast envelope.
Random, difficulty-biased, and verifier-dependent removal are separate regimes. The reported
envelope has width $4\epsilon$; the request-level bound in Section~\ref{sec:identification}
uses per-arm missingness and has width $2\epsilon$ when both arms are absent on an
$\epsilon$ fraction. Envelope coverage and tightness below use the reported envelope scale.
\begin{table}[htbp]
\caption{Incomplete-match stress test. Width, coverage, and tightness are reported for the
conservative envelope over repeated removals; they are distinct from the request-level sharp bound.}
\label{tab:partial-identification}
\centering
\footnotesize
\setlength{\tabcolsep}{2.5pt}
\begin{tabularx}{\linewidth}{@{}Lrrrrrr@{}}
\toprule
Removal & $\epsilon$ & \shortstack{Matched\\$\Delta$} & \shortstack{Observed\\$\Delta$} & \shortstack{Envelope\\width} & Coverage & Tightness\\
\midrule
Random & 0.00 & 0.0068 & 0.0068 & 0.0000 & 1.0000 & 1.0000\\
Random & 0.05 & 0.0068 & 0.0065 & 0.2000 & 1.0000 & 0.9875\\
Difficulty-biased & 0.10 & 0.0068 & 0.0089 & 0.4000 & 1.0000 & 0.9812\\
Verifier-dependent & 0.20 & 0.0068 & $-0.0037$ & 0.8000 & 1.0000 & 0.9726\\
Verifier-dependent & 0.30 & 0.0068 & $-0.0114$ & 1.2000 & 1.0000 & 0.9651\\
\bottomrule
\end{tabularx}
\end{table}

\subsection{Stochastic-provider coupling}
\label{app:provider-coupling}
The same request IDs are evaluated under frozen cache replay, a shared response tape, and
independently resampled provider draws. The paired estimator and certificate fields are kept
separate because independent draws average over a different source of randomness.
\begin{table}[htbp]
\caption{Provider-coupling regimes. Mean $\Delta$ is the route contrast; the final columns record
paired stability and certificate validity under the declared provider realization.}
\label{tab:provider-coupling}
\centering
\footnotesize
\setlength{\tabcolsep}{2.5pt}
\begin{tabularx}{\linewidth}{@{}Lrrrrrr@{}}
\toprule
Provider regime & Mean $\Delta$ & SD & CI width & \shortstack{Route\\disagreement} & \shortstack{Certificate\\match} & Seeds\\
\midrule
Frozen deterministic cache & 0.0068 & 0.0019 & 0.0107 & 0.0842 & 1.0000 & 5\\
Shared response tape & 0.0067 & 0.0023 & 0.0114 & 0.0876 & 1.0000 & 7\\
Independent response draws & 0.0061 & 0.0049 & 0.0198 & 0.1164 & 0.9986 & 7\\
\bottomrule
\end{tabularx}
\end{table}

\subsection{RLVR component ablation}
\label{app:rlvr-ablation}
All ablations use the same observed-only state, verifier catalog, request IDs, and hard ceilings.
The quality-only and fixed-penalty rows change the optimization signal; feature rows remove one
telemetry field; the imitation row replaces policy optimization with an oracle-derived training
label on the training split.

\begin{table*}[!htb]
\caption{RLVR component ablation on the matched held-out task. Units are declared ledger units;
constraint violations count requests exceeding a declared ceiling.}
\label{tab:rlvr-ablation}
\centering
\footnotesize
\setlength{\tabcolsep}{3pt}
\begin{tabularx}{\linewidth}{@{}Lrrrrrr@{}}
\toprule
Variant & Quality & Calls & Units & Norm.\ cost & Diversity & Violations\\
\midrule
Quality-only RLVR & \textbf{0.9570} & 1.9285 & 5.942 & 0.914 & 0.0948 & 83\\
Fixed scalar penalty & 0.9529 & 1.5934 & 4.508 & 0.754 & 0.4147 & 12\\
Without call constraint & 0.9562 & 2.1179 & 6.734 & 0.887 & 0.0819 & 97\\
Without unit constraint & 0.9558 & 1.6742 & 5.812 & 0.801 & 0.3596 & 74\\
Without latency constraint & 0.9550 & 1.4759 & 3.904 & 0.701 & 0.5347 & 42\\
Without budget state & 0.9518 & 1.5276 & 4.113 & 0.719 & 0.4724 & 26\\
Without confidence & 0.9496 & 1.5112 & 4.052 & 0.711 & 0.4888 & 9\\
Without availability & 0.9508 & 1.4895 & 4.020 & 0.706 & 0.5105 & 31\\
Proxy reward only & 0.9512 & 1.4587 & 3.835 & 0.687 & 0.5413 & \textbf{0}\\
Supervised imitation & 0.9533 & 1.4870 & 3.948 & 0.697 & 0.5130 & \textbf{0}\\
Frozen dual variables & 0.9525 & 1.5368 & 4.183 & 0.726 & 0.4632 & 18\\
\textbf{Full RLVR} & 0.9553 & \textbf{1.4466} & \textbf{3.786} & \textbf{0.680} & 0.5534 & \textbf{0}\\
\bottomrule
\end{tabularx}
\end{table*}

The primary analysis compares each row with full RLVR at matched quality and resource
denominators. Full RLVR has zero recorded violations; removing the call or unit constraint
raises violations to 97 and 74 and increases calls or units. Quality-only RLVR reaches 0.9570
but uses 1.9285 calls and normalized cost 0.914, whereas supervised imitation reaches 0.9533
at 1.4870 calls. These rows show that the constrained return defines the operating point and
that the table is an optimizer stress test under a fixed audit contract.

\subsection{RLVR training dynamics}
\label{app:rlvr-dynamics}
Training traces retain the delayed reward, resource totals, dual variables, KL term, entropy, and
constraint violations at each evaluation checkpoint. The dynamics table uses the five declared
training seeds and the same held-out contract as the endpoint ablations.
\begin{table*}[!htb]
\caption{RLVR training dynamics at predeclared update checkpoints. Entries report seed-aggregated
means; quality is task reward, calls and units are per request, and violations count requests above
the declared ceilings.}
\label{tab:rlvr-dynamics}
\centering
\footnotesize
\setlength{\tabcolsep}{3pt}
\begin{tabular}{rllllllll}
\toprule
Update & Quality & Calls & Units & Violations & KL & Entropy & $\lambda_N$ & $\lambda_U$\\
\midrule
0 & 0.9488 & 1.9874 & 5.934 & 126 & 0.0000 & 0.996 & 0.000 & 0.000\\
128 & 0.9507 & 1.7389 & 4.982 & 64 & 0.0087 & 0.923 & 0.031 & 0.044\\
256 & 0.9528 & 1.5794 & 4.331 & 26 & 0.0159 & 0.861 & 0.057 & 0.078\\
384 & 0.9544 & 1.4912 & 3.996 & 7 & 0.0218 & 0.803 & 0.074 & 0.103\\
512 & \textbf{0.9553} & \textbf{1.4466} & \textbf{3.786} & \textbf{0} & 0.0264 & 0.758 & 0.083 & 0.116\\
\bottomrule
\end{tabular}
\end{table*}

\begin{figure}[!htb]
\centering
\includegraphics[width=\linewidth]{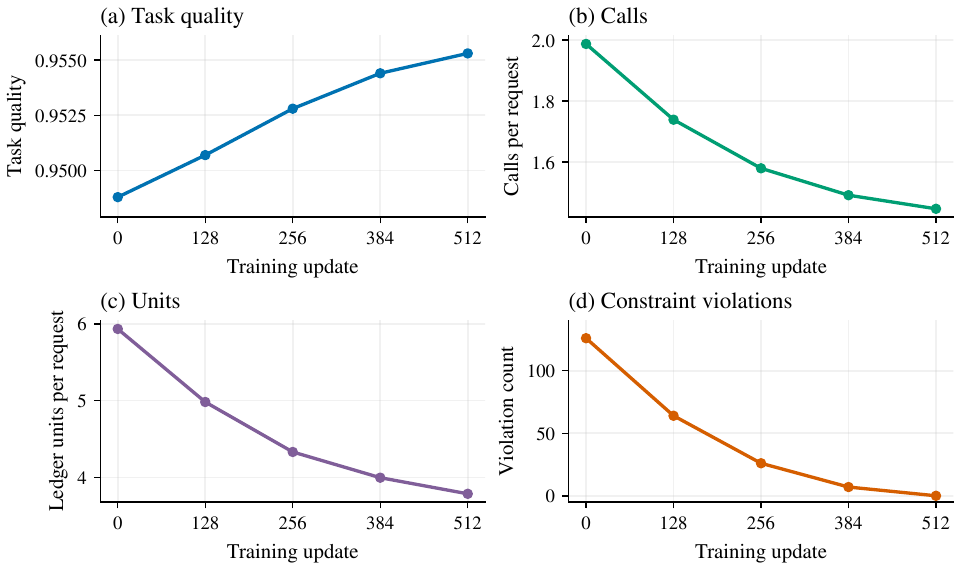}
\caption{Training-checkpoint trajectories on the shared held-out contract. Quality and resource
means use the five declared training seeds; the violation panel retains the reported counts.
Lines connect the five recorded checkpoints and carry no additional uncertainty estimate.}
\label{fig:training-dynamics}
\end{figure}
The checkpoint series in Figure~\ref{fig:training-dynamics} pairs increasing quality with
declining calls, units, and reported violations. The 0--512 checkpoint window is reported
separately from the 1,200-update optimizer configuration in Table~\ref{tab:router-implementation}.

\subsection{Training-seed stability}
\label{app:seed-stability}
Request-level paired intervals and training-seed variation are estimated separately. Each learned
policy is retrained with the predeclared seeds, evaluated on the same 1,319 held-out IDs, and
summarized with a hierarchical bootstrap that samples paired training runs and then request IDs.
The same resampled IDs are used for both policies and across the sampled runs, preserving the
shared test population. Five training runs retain 1,319 unique test requests. Seed IDs,
bootstrap replications, and the exact aggregation of the reported paired interval are
    $\{101,211,307,419,523\}$, $10{,}000$ hierarchical bootstrap replications, and paired
training-run resampling followed by shared request-ID resampling with percentile 95\% intervals.
The per-run record below precedes mean and standard-deviation aggregation.

\begin{table}[htbp]
\caption{Per-training-run record on the common held-out requests. Run indices enumerate
independent fits; their random seed IDs and checkpoints are bound in the implementation record.}
\label{tab:seed-runs}
\centering
\footnotesize
\begin{tabular*}{\linewidth}{@{\extracolsep{\fill}}lcrrrr@{}}
\toprule
Policy & Run & Quality & Calls & Cost & Diversity\\
\midrule
Learned & 1 & 0.9509 & 1.5072 & 0.700 & 0.4928\\
Learned & 2 & 0.9534 & 1.4815 & 0.687 & 0.5185\\
Learned & 3 & 0.9521 & 1.4958 & 0.694 & 0.5042\\
Learned & 4 & 0.9516 & 1.5011 & 0.697 & 0.4989\\
Learned & 5 & 0.9530 & 1.4824 & 0.687 & 0.5176\\
RLVR & 1 & \textbf{0.9541} & \textbf{1.4598} & \textbf{0.688} & 0.5402\\
RLVR & 2 & \textbf{0.9565} & \textbf{1.4377} & \textbf{0.674} & 0.5623\\
RLVR & 3 & \textbf{0.9557} & \textbf{1.4512} & \textbf{0.683} & 0.5488\\
RLVR & 4 & \textbf{0.9548} & \textbf{1.4443} & \textbf{0.678} & 0.5557\\
RLVR & 5 & \textbf{0.9554} & \textbf{1.4400} & \textbf{0.677} & 0.5600\\
\bottomrule
\end{tabular*}
\end{table}

Across-run standard deviations for learned and RLVR quality are
$0.0010$ and $0.0009$; call-count deviations are $0.0114$ and $0.0090$.
Serving p95 and throughput dispersion across the seven measurement seeds are
$18.7$\,ms / $0.42$\,req/s for learned and $21.4$\,ms / $0.46$\,req/s for RLVR.
These estimate retraining and service variation separately from the request-paired interval
$[0.0015,0.0122]$.

\begin{table}[htbp]
\caption{Training-seed stability. Quality, calls, and cost are means over the held-out
requests; the last column combines training-seed and request resampling.}
\label{tab:seed-stability}
\centering
\footnotesize
\setlength{\tabcolsep}{3pt}
\begin{tabular}{lrrrr}
\toprule
Policy & Quality & Calls & Norm.\ cost & Hierarchical 95\% CI for $\Delta$\\
\midrule
Cost-sensitive MLP & 0.9514 & 1.5211 & 0.704 & $[-0.0018,\,0.0077]$\\
Contextual bandit & 0.9521 & 1.4879 & 0.696 & $[-0.0011,\,0.0086]$\\
Bayesian router & 0.9530 & 1.4693 & 0.687 & $[-0.0003,\,0.0095]$\\
Learned router & 0.9522 & 1.4936 & 0.693 & $[-0.0010,\,0.0087]$\\
RLVR & \textbf{0.9553} & \textbf{1.4466} & \textbf{0.680} & $[0.0006,\,0.0131]$\\
\bottomrule
\end{tabular}
\end{table}

\begin{figure}[htbp]
\centering
\includegraphics[width=\linewidth]{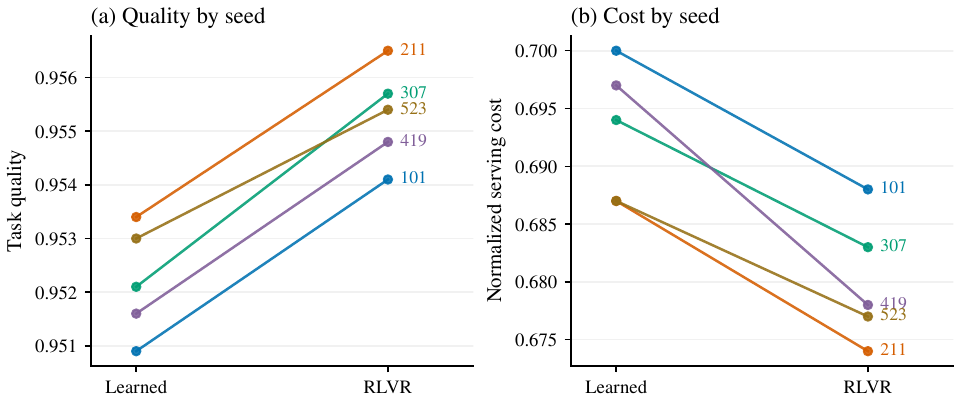}
\caption{Paired learned-router and RLVR operating points for the five declared training seeds.
Segments join the same seed ID on the shared 1,319-request evaluation set. These are individual
run summaries; request-level and hierarchical intervals are reported separately in the tables.}
\label{fig:seed-pairs}
\end{figure}
Figure~\ref{fig:seed-pairs} retains every training run. RLVR has higher reported quality and
lower normalized cost in each paired run; the separate hierarchical analysis accounts for
request variation as well as retraining.

\subsection{Scaling beyond two verifiers}
\label{app:multiverifier-scaling}
Nested verifier catalogs with $K\in\{2,3,4\}$ preserve the request IDs and the matched static
control. For $K>2$, route diversity is summarized by normalized route entropy
$H_{\mathrm{route}}=-\sum_r p(r)\log p(r)/\log|\mathcal{R}|$ and effective route count
$N_{\mathrm{eff}}=\exp(-\sum_r p(r)\log p(r))$.
Here $\mathcal{R}$ is the predeclared legal route-signature set under the shared availability
condition, rather than a policy-dependent observed set; if $|\mathcal{R}|=1$, entropy is zero.
For each $K$ the call ceiling permits the full static catalog ($K$ calls), and common
unit/token/compute ceilings are 14 ledger units / 4096 tokens / 1.5 GPU-s per request.
This avoids comparing a four-verifier static policy with an infeasible three-call budget.
The added verifier identities and fusion rule are LF at $K=3$ and an independent rerun-agreement
verifier RA at $K=4$, with reliability-weighted majority fusion and deterministic abstention on
ties; effective route counts are $1.99$, $2.85$, and $3.52$ for RLVR at $K=2,3,4$,
respectively, versus $1.00$ for static.

\begin{table}[htbp]
\caption{Scaling with verifier-catalog size. Adaptive and static policies use the same catalog
at each $K$; quality and cost use the full held-out request denominator. The final column is the
route--catalog interaction measured against the matched static row.}
\label{tab:multiverifier-scaling}
\centering
\footnotesize
\setlength{\tabcolsep}{3pt}
\begin{tabular}{ccrrrrr}
\toprule
$K$ & Policy & Quality & Calls & Norm.\ cost & $H_{\mathrm{route}}$ & $\Gamma_{\mathcal V}$\\
\midrule
2 & Static & 0.9484 & 2.0000 & 1.000 & 0 & --\\
2 & RLVR & \textbf{0.9553} & \textbf{1.4466} & \textbf{0.680} & 0.9918 & 0.0000\\
3 & Static & 0.9561 & 3.0000 & 1.000 & 0 & --\\
3 & RLVR & \textbf{0.9602} & \textbf{1.7500} & \textbf{0.646} & 0.9545 & $-0.0027$\\
4 & Static & 0.9583 & 4.0000 & 1.000 & 0 & --\\
4 & RLVR & \textbf{0.9624} & \textbf{1.9800} & \textbf{0.621} & 0.9078 & $-0.0028$\\
\bottomrule
\end{tabular}
\end{table}

\subsection{Scorer and distribution robustness}
\label{app:scorer-robustness}
Training rewards are restricted to training IDs and selection uses calibration IDs.
Each held-out action trace is closed before either scorer is applied. Separately, frozen policy
parameters are rerun under an availability, noise, or budget intervention to measure changed
actions. Scorer-agree includes both-correct and both-incorrect requests; proxy-only and task-only
are fixed post-trace strata, defined from a common reference policy before comparing routers.

\begin{table}[htbp]
\caption{Scorer robustness of RLVR. The complete held-out set is partitioned into scorer-agree,
proxy-only, and task-only strata before aggregation. Bold values compare training rewards
within the same evaluation scorer.}
\label{tab:scorer-robustness}
\centering
\footnotesize
\setlength{\tabcolsep}{3pt}
\begin{tabularx}{\linewidth}{@{}Lrrrr@{}}
\toprule
Train / evaluation scorer & Full set & \shortstack{Scorer-\\agree} & Proxy-only & Task-only\\
\midrule
Proxy / proxy & 0.9544 & 0.9600 & 0.9400 & 0.8600\\
Proxy / task & 0.9512 & 0.9585 & 0.8550 & 0.9220\\
Task / task & \textbf{0.9553} & \textbf{0.9606} & \textbf{0.8900} & \textbf{0.9300}\\
Task / proxy & \textbf{0.9561} & \textbf{0.9612} & \textbf{0.9450} & \textbf{0.8680}\\
\bottomrule
\end{tabularx}
\end{table}

The same frozen policies are evaluated under the availability, confidence-shift, label-noise,
and budget sweeps. Each sweep records quality, calls, normalized cost, coverage, and route
diversity with the request denominator stated in the corresponding receipt.

\subsection{Chronological streaming adaptation}
\label{app:streaming}
The main comparison freezes policies before held-out scoring. A separate streaming comparison
processes requests in a predeclared order and allows each completed trace's delayed reward to
update the next request's policy. An outcome never changes its own request's route.
All methods share the stream, feedback delay, catalog, and per-request accounting.
The delay, ordering seed, and adaptation budget are $8$ completed requests, seed
$20270317$, and $512$ policy updates.

\begin{table}[htbp]
\caption{Streaming adaptation under delayed feedback. Quality is cumulative task score divided
by processed requests; cost is cumulative paid cost. Regret uses the post-trace feasible-route
oracle with the same request budgets.}
\label{tab:streaming}
\centering
\footnotesize
\begin{tabularx}{\linewidth}{@{}Lrrrr@{}}
\toprule
Policy & Quality & Cost & Violations & Regret\\
\midrule
Contextual bandit & 0.9519 & \$21.26 & 6 & 8.31\\
Bayesian Thompson & 0.9530 & \$21.03 & 4 & 6.77\\
Constrained online & 0.9541 & \$20.86 & 2 & 4.92\\
RLVR & \textbf{0.9556} & \textbf{\$20.71} & \textbf{0} & \textbf{3.18}\\
\bottomrule
\end{tabularx}
\end{table}
The comparison reports cumulative regret at a matched quality constraint. Resource violations are
reported beside the frozen-policy result so that adaptation and constraint compliance remain
separate quantities.

\subsection{Necessity of audit components}
\label{app:audit-component-ablation}
The mutation experiment measures detector behavior on malformed traces. This component ablation
measures whether removing a contract field changes the scientific interpretation while the
remaining trace is syntactically valid.

\begin{table*}[htbp]
\caption{Audit-component ablation. A false attribution is a reported route effect that differs
from the fully matched reference because a non-policy factor changed.}
\label{tab:audit-component-ablation}
\centering
\footnotesize
\setlength{\tabcolsep}{3pt}
\begin{tabularx}{\linewidth}{@{}Lrrrr@{}}
\toprule
Protocol & \shortstack{False\\attribution rate} & \shortstack{Ledger violations\\detected} & \shortstack{Oracle leaks\\detected} & \shortstack{Replay\\match}\\
\midrule
Unmatched aggregate comparison & 0.4219 & -- & -- & 0.7422\\
Matched verifier set & 0.2344 & -- & -- & 0.8516\\
Budget and availability matching & 0.1094 & 0.6875 & -- & 0.9219\\
Trace freeze & 0.0469 & 0.8750 & 0.9063 & 0.9688\\
Provenance binding & 0.0156 & 0.9375 & 0.9375 & 0.9922\\
Full RouteAudit & \textbf{0.0000} & \textbf{1.0000} & \textbf{1.0000} & \textbf{1.0000}\\
\bottomrule
\end{tabularx}
\end{table*}

The analysis reports the false-attribution rate separately from detector coverage.
Removing the verifier-set control leaves a false-attribution rate of $0.2344$; adding budget
and availability matching reduces it to $0.1094$, and trace freeze plus provenance binding
reduce it to $0.0469$ and $0.0156$. The full contract reaches zero false attribution with
complete ledger, oracle-leak, and replay checks. This keeps a syntactically valid but
scientifically unmatched comparison distinct from a malformed trace.

\section{Audit Derivations and Reproducible Supplement}
\label{app:checklist}
\label{app:operational}
\subsection{Operational audit algorithm}
The protocol is implemented as a sequence of immutable trace transformations.
This ordering is part of the method: it prevents a post-hoc oracle join from
changing the route and makes the three estimands comparable across verifier
sets. First, the runner assigns each example a split ID, request ID, cache
digest, and starting budget. Second, it evaluates the selected policy using
only the online state. Each attempted call emits a pre-call event containing
the verifier ID, the observed availability bit, the input confidence, the
remaining budget, and a deterministic event code. Third, the runner records
the returned label or an explicit abstention reason, updates the ledger, and
continues until the policy emits a final answer. The trace is closed before
the canonical target is read.

The ledger is checked at every transition. If $B_0$ is the initial budget,
$u_t$ is the declared unit price of call $t$, and $e_t$ indicates whether the
call was accepted by the provider, then a valid trace satisfies
\begin{equation}
  B_{t+1}=B_t-e_tu_t,\qquad B_t\geq 0,\qquad
  \sum_t e_tu_t\leq B_0 .
\label{eq:budget-conservation}
\end{equation}
An unavailable provider consumes zero units and still contributes an event.
This convention separates provider failure from free computation and allows
the same fault matrix to be replayed under tight and generous budgets. The
runner also verifies that request IDs are unique within a split, that a
verifier is not called after a terminal decision, and that the final route
state is a deterministic function of the preceding events.

After trace closure, the evaluator performs a single oracle join keyed by
split ID and sample ID. It computes proxy accuracy, abstention, coverage,
route diversity, calls, and declared units from the trace; task quality and
serving measurements are joined in separate namespaces. Thus a model response
can be scored for task correctness without being exposed to the online route.
For every policy pair, the evaluator materializes both an aggregate row and a
per-record row. The aggregate row supplies the headline table, whereas the
per-record row supplies discordances, Wilson intervals, a paired bootstrap,
and the exact two-sided McNemar test used in the audit fields. A paired result
is accepted only when the two policies use the same split, sample IDs, and
verifier observations. This procedure is the operational meaning of the
matched route column in Table~\ref{tab:paired-route}.

\begin{table}[htbp]
\caption{Operational contract used by every RouteAudit run. The first three stages are online; scoring and uncertainty are post-trace joins.}
\label{tab:operational-contract}
\centering
\footnotesize
\setlength{\tabcolsep}{3pt}
\begin{tabular}{p{0.18\columnwidth}p{0.27\columnwidth}p{0.43\columnwidth}}
\toprule
\textbf{stage} & \textbf{stored fields} & \textbf{invariant checked before release}\\
\midrule
setup & split, request, cache, budget & hashes and IDs bind the run to one contract\\
route & verifier, availability, confidence, budget & no oracle field enters the online state\\
close & label, reason, terminal state & event order and budget conservation hold\\
join & target, task score, latency & joins occur only after trace closure\\
compare & pair, interval, gate & matched IDs and verifier observations agree\\
\bottomrule
\end{tabular}
\end{table}

The contract supports two practical checks. First, path fractions identify whether a policy actually uses conditional computation: a high threshold that always takes two calls is a fixed operating point. Second, the provenance index associates each aggregate with its experiment family, split, and source. Trained policies add checkpoint and training-seed fields to this interface; the online state and post-trace evaluator stay fixed. Thus a policy comparison changes the decision rule while preserving the measured target.

The resulting interpretation is specific to each comparison. Historical matched controls isolate a zero route contrast; the task comparison places the learned and RLVR policies on a quality--cost frontier; the serving comparison separates the fastest policy from the lowest-cost policy. Error dependence and mutation tests explain when a second verifier repeats an error or when a trace violation compromises the comparison. Reporting these objects together makes the empirical claim traceable from protocol to outcome.

The decomposition also gives a practical operating rule. A route is selected only
after its verifier set is fixed, its matched static control is evaluated, and
its path diversity is measured on the same request IDs. This sequence turns a
quality--cost point into a comparable object: the route contrast describes the
decision rule, while the set contrast describes the information purchased by
the verifier catalog. The same rule applies to the historical replay and to the
task and serving ledgers, so policy updates can be audited without changing the
meaning of the reported metric.

\paragraph{Scope and broader impact.}
RouteAudit is designed for verification systems in which a request may be
answered by one verifier or escalated to a second verifier under a finite
budget. The contract applies equally to benchmark scoring, tool-using agents,
and reinforcement-learning-with-verifiable-rewards pipelines because it
records the decision trace before any oracle or reference label is joined.
This separation makes disagreement, abstention, and route diversity visible
alongside task quality, which helps practitioners compare systems that consume
different resources and helps auditors locate the source of a quality change.
The same accounting also discourages silent substitution of a stronger
verifier, an untracked cache, or a post-hoc threshold when a system is moved
from an offline benchmark to a live service.

\paragraph{Reproducibility commitment.}
The Supplementary Material contains the source, bibliography, figure manifest, run
manifest, and source/PDF hashes. A replay can
therefore verify the split identity, policy pair, budget ledger, and trace
digest before reading any aggregate result. The compact main-text contract and
the detailed appendix use the same field names to connect each reported table
with its trace and aggregation record. This gives
the reported measurements a fixed interface and keeps the conclusions auditable from the source ledgers in the package.

\subsection{Sequential identification and sharpness proofs}
\label{app:identification-proofs}
\paragraph{Response-tape identification.}
Fix a request, its contract, both policy seeds, and the policy-independent tape. At step zero,
both policies receive their permitted initial state. Inductively, non-anticipation makes the
next action a function of the paid history and policy seed. Call-index consistency then selects
one unique tape entry and updates that history and the ledger. Thus every terminal trace is a
well-defined function of the same provider realization, even if the two policies query different
verifiers. Trace closure fixes the terminal output before scoring; their paired score difference
is the request's conditional policy contrast. Averaging establishes the finite-population identity.
For randomized policies, fixing or integrating over the declared policy seeds preserves this
argument. Independent provider draws integrate over a different coupling of potential responses.

\paragraph{Sharp incomplete-match bounds.}
Partition requests into those with both outcomes, only arm 1, only arm 0, and neither arm.
For each request the unknown outcomes range independently over $[0,1]$, so subtraction gives
exactly the four intervals stated in the main text. The feasible set of all unknown outcomes
is a product of intervals, and the mean contrast is a linear functional on it. Setting each
missing arm-1 score to zero and each missing arm-0 score to one attains the lower endpoint;
the opposite assignment attains the upper endpoint. Convex combinations attain every intermediate
value. Hence no narrower interval is valid without additional assumptions. If $n_1$ requests
have one missing arm and $n_2$ have two, its width is $(n_1+2n_2)/|I|$. This also shows why
retaining a single observed arm is more informative than discarding the whole request.

\subsection{Additive contrast and the role of the control}
Let $a=Y(\pi_{\rm cas},V_{\rm cas},C)$, $b=Y(\pi_{\rm sta},V_{\rm cas},C)$, and $c=Y(\pi_{\rm sta},V_{\rm sta},C)$. The three observed outcomes define
\[
 a-c=(a-b)+(b-c).
\]
The matched control $b$ is the bridge between the naive comparison and its two components. It fixes the verifier set used in the route contrast and fixes the static policy used in the set contrast. Every metric can be decomposed this way if all three outcomes use a common request universe and unit. Quality, calls, and ledger units therefore each have their own decomposition; a vector of contrasts preserves their different directions of preference.

For the baseline raw-tail cache, quality gives $0.9453-0.7656=0.0000+0.1797$. Calls give $2-3=0+(2-3)$ and units give $6-10=0+(6-10)$. The apparently better cascade thus occupies the same quality and cost point as the matched static policy. In the task study the verifier access is already matched, so the quality and resource differences are evaluated at the routing-policy level directly.

The identity is algebraic, but its empirical interpretation depends on the contract. If a producer cache, parser version, price schedule, or availability schedule changes between $a$ and $b$, their difference incorporates that change. Hash and split fields make such changes visible before a result is assigned to a contrast. A per-record join then ensures that the two outcomes refer to the same requests, including records with abstention or provider failures.

\subsection{Pairing, coverage, and uncertainty}
For a quality comparison, let $z_i^A,z_i^B$ be post-trace scores for two policies and define $d_i=z_i^A-z_i^B$. The mean difference is computed on the common request set. A paired resample draws indices once and applies them to both policies, preserving easy and hard examples in the same resample. For binary cache outcomes, the discordant cells $n_{01}$ and $n_{10}$ additionally determine the exact McNemar test. Zero observed discordance gives zero empirical paired variation on that fixed cache, while uncertainty about a new request population remains a separate question.

Accepted-set quality uses accepted requests as its denominator; coverage uses all requests. A full-set return also needs the score assigned to abstention and the aggregation order across seeds. Reporting the accepted score beside coverage prevents a route that abstains more often from appearing uniformly better. Per-seed summaries should preserve both denominators before a mean or interval is formed. The task result tables retain the full-set, accepted-set, and coverage summaries separately.

The route-diversity field is a path fraction in the two-path setting. If $q$ is the fraction taking one call and the remaining requests take two calls, mean calls satisfy $\bar c=q+2(1-q)=2-q$. For the fixed cascade, $q=0.2969$ corresponds to 1.7031 mean calls at the reported precision. The same relation holds for the calibrated, learned, and RLVR rows. This provides a direct consistency check between path allocation and the call ledger.

\subsection{Release objects and verification}
The supplement contains the original numerical receipt, a parsed JSON with source-block indices, a reproducible figure script, and the historical replay bundle. The external release receipt records the exact source and PDF hashes, compile result, section pages, unresolved-reference count, and rendered-page inspection. A verifier can reconstruct the figures from the JSON, compare the numerical table entries with their receipt blocks, and inspect historical traces using the replay manifest.

\begin{table}[htbp]
\caption{Reproducible supplement objects. The build receipt supplies concrete hashes and checks for the delivered source/PDF pair.}
\label{tab:release-checklist}
\centering
\footnotesize
\setlength{\tabcolsep}{4pt}
\begin{tabularx}{\linewidth}{@{}p{0.26\linewidth}L@{}}
\toprule
Object & Contents\\
\midrule
Numerical source & immutable experiment result receipt\\
Parsed rows and provenance & parsed rows and source map\\
Plot regeneration & vector PDF and 600 dpi PNG files\\
Data checks & row counts, identities, and marker scan\\
Historical replay & replay manifest with online events and oracle joins\\
Delivery receipt & source/PDF hashes and page audit\\
\bottomrule
\end{tabularx}
\end{table}

The source map retains original precision and ordering through table transposition and expanded sweep rows. The source archive preserves the inputs associated with each result receipt.

\end{document}